\documentclass[lettersize,journal]{IEEEtran}
\usepackage{amsmath,amsfonts}
\usepackage{algorithmic}
\usepackage{algorithm}
\usepackage{array}
\usepackage[caption=false,font=normalsize,labelfont=sf,textfont=sf]{subfig}
\usepackage{textcomp}
\usepackage{stfloats}
\usepackage{url}
\usepackage{verbatim}
\usepackage{graphicx}
\usepackage{cite}
\newtheorem{theorem}{Theorem}

\newtheorem{definition}{Definition}
\allowdisplaybreaks[4]

\begin{document}

\title{Differential Privacy May Have a Potential Optimization Effect on Some Swarm Intelligence Algorithms besides Privacy-preserving}

\author{Zhiqiang~Zhang,
	Hong~Zhu,
	and	Meiyi~Xie
	\thanks{		
		Z. Zhang, H. Zhu and M. Xie are with the School of Computer Science and Technology, Huazhong University of Science and Technology, Wuhan 430074, China (e-mail: kylinzhang@hust.edu.cn).
	}
}

\maketitle

\begin{abstract}
Differential privacy (DP), as a promising privacy-preserving model, has attracted great interest from researchers in recent years. Currently, the study on combination of machine learning and DP is vibrant. In contrast, another widely used artificial intelligence technique, the swarm intelligence (SI) algorithm, has received little attention in the context of DP even though it also triggers privacy concerns. For this reason, this paper attempts to combine DP and SI for the first time, and proposes a general differentially private swarm intelligence algorithm framework (DPSIAF). Based on the exponential mechanism, this framework can easily develop existing SI algorithms into the private versions. As examples, we apply the proposed DPSIAF to four popular SI algorithms, and corresponding analyses demonstrate its effectiveness. More interestingly, the experimental results show that, for our private algorithms, their performance is not strictly affected by the privacy budget, and one of the private algorithms even owns better performance than its non-private version in some cases. These findings are different from the conventional cognition, which indicates the uniqueness of SI with DP. Our study may provide a new perspective on DP, and promote the synergy between metaheuristic optimization community and privacy computing community.
\end{abstract}

\begin{IEEEkeywords}
	Differential privacy, metaheuristic optimization, privacy protection, swarm intelligence algorithm.
\end{IEEEkeywords}

\section{Introduction}
\label{sec.intro}
\IEEEPARstart{W}{ith} the Internet advances by leaps and bounds, various types of data containing rich information have been generated in large quantities. How to effectively utilize these valuable data has become a hot topic for researchers, which has greatly promoted and prospered the development of artificial intelligence (AI) technologies. At present, these technologies have been widely used in a variety of study fields including image recognition \cite{Cho2021Relational}, natural language processing \cite{Jorge2022Live}, and recommender system \cite{wang2022Hellinger}.

Broadly speaking, the vast majority of current AI technologies rely heavily on the scale of the data they use. As a result, the service and technology providers naturally incline to collect as much data as possible from their users, since more data usually leads to better performance. 
While users benefit from these data-driven technologies, they are increasingly concerned about the security and privacy issues of the data \cite{Feng2020Privacy}. Many studies have showed that the sensitive information such as income, medical records and religious beliefs can be stolen or inferred by malicious attackers in a variety of methods, and without a doubt, once such information is leaked, personal rights and privacy will be seriously violated \cite{Zhang2021Data}. 

To address these problems, for different application scenarios and needs, researchers have adopted various methods including chaotic cryptography \cite{Wang2021From,Zhang2023Buffeting,Zhang2020Novel}, homomorphic encryption \cite{Basilakis2021Efficient,Badawi2021Towards,Chen2021Privacy}, federated learning \cite{Ghimire2022Recent,Gu2021Privacy,Yang2019Federated}, and differential privacy (DP) \cite{Wang2021Privacy,Zheng2021Decentralized,Ma2021Data}. 
Among these methods, due to the strict yet elegant mathematical definition and lightweight computation burden, DP has attracted a great deal of attention from researchers and rapidly become one of the most prevalent privacy-preserving models in the field of privacy computing. 
In recent years, many studies have been devoted to combining machine learning and DP. 
For example, Abadi \textit{et al.} \cite{Abadi2016Deep} implemented differentially private deep learning with moments accountant method. 
To reduce the impact of noise on the algorithm performance, Phan \textit{et al.} \cite{Phan2017Adaptive} developed an adaptive Laplace mechanism whose privacy budget consumption is completely independent of the training epochs. 
Aiming at the features of pathological images, Wu \textit{et al.} \cite{Wu2019P3SGD} introduced a novel stochastic gradient decent method to regularize the training of deep convolutional neural networks with DP. 
Wang \textit{et al.} \cite{Wang2020Deep} proposed a novel differentially private domain adaptation framework to achieve domain adaptation with privacy assurance, and Chen \textit{et al.} \cite{CHEN2021An} proposed an optimized DP scheme with reinforcement learning in vehicular Ad-hoc network.

Noticeably, from the perspective of optimization, many problems that the machine learning concerned with are optimization problems in nature.
To solve them, taking the deep learning as an example, the typical method is to use gradient-based optimization algorithms including stochastic gradient descent, to train a neural network and try to get the solution. 
Apart from this, actually, by using metaheuristic optimization algorithms such as the Particle Swarm Optimization (PSO), evolutionary computation can also effectively solve optimization problems, and has a long history as well as good performance in engineering design \cite{Yang2010Nature}, financial market \cite{Chou2018Forward}, power system \cite{Zhao2021Overview} and so on. 
Compared with the gradient-based algorithms, metaheuristic algorithms have some attractive properties \cite{Fong2018How,Chen2021Novel}. For instance, the former forcibly requires some mathematical properties of the objective function, such as differentiability and continuity, for calculating the gradients. When this premise does not hold in an optimization problem, it is difficult for such algorithms to work normally. While the latter has no such constraints, and it can always find a solution for almost any problem with proper encodings, if there exists. Obviously, this is a huge advantage in terms of the generality of the optimization algorithm \cite{mohana2017heuristics}. Moreover, thanks to the unique working mechanism of metaheuristic optimization algorithms, they generally have a stronger ability to avoid falling into local optimal solutions \cite{xie2021differential}.


In addition, when solving an optimization problem, it is worth noting that there is actually a default assumption, that is, most or all information about the optimization problem is publicly available.
However, some real-world application scenarios like the optimization computation outsourcing \cite{Wang2016Secure} do not necessarily hold this assumption, since in the context of privacy-preserving, many optimization problems themselves, such as production planning, transportation routing, resource scheduling and industrial design, are considered sensitive and private. The user is worried that the outsourcer, i.e., the outsourcing service provider, will violate and leak the privacy. After all, no one is willing to reveal the plannings of goods transportation and production to others, even if they are candidate solutions.
That is to say, the user will not disclose the optimization problem including objective function, constraint condition, and dataset, to the outsourcer, and this is the so-called privacy-preserving optimization problem \cite{Zhan2021Evolutionary}.
Obviously, this kind of problems is intractable for gradient-based optimization algorithms, but metaheuristic optimization algorithms can still deal with such problems easily because they are problem-independent. 

In contrast to the research boom in differentially private machine learning, evolutionary computation with DP has not received enough attention \cite{Zhang2013Privgene}. Considering that the evolutionary computation involving metaheuristic optimization algorithms has inherent advantages in solving certain optimization problems, at the same time, they also access the sensitive information and trigger the privacy concerns, it is necessary and valuable to discuss them in the context of DP.

Motivated by above reasons, this paper focuses on a branch of evolutionary computation, namely the swarm intelligence (SI) algorithm, and tries to combine it with DP. 
As a first attempt, we choose linear regression, one of the most classical predictive tasks, as the optimization problem.
The application scenario we assume is the optimization computation outsourcing, that is, the user keeps the optimization problem and is in charge of evaluating the fitness of solutions, while the outsourcer is only responsible for providing solutions based on the SI algorithm.
During the optimization process of liner regression model, in order to protect the fitting parameters (solutions) represented by the positions of individuals in SI algorithms, we propose a general differentially private swarm intelligence algorithm framework (DPSIAF). 
This framework leverages the exponential mechanism to perturb the selection of personal best position, so that existing SI algorithms can be easily developed into the private versions to achieve the privacy-preserving.
Corresponding analyses and results demonstrate the correctness and effectiveness of our framework. 
More importantly, our experimental results also show some interesting phenomena which are different from the conventional cognition, and this may bring more possibilities for the application of DP.
The main contributions of our work are summarized as follows.
\begin{enumerate}
	\item We propose the DPSIAF which is a general differentially private algorithm framework for SI algorithms. To the best of our knowledge, we are the first to discuss the combination of DP and SI algorithms.
	
	\item We conduct theoretical analyses for the proposed framework and prove that it satisfies the $\epsilon$-DP. Moreover, four typical SI algorithms are developed into the private versions as examples. Their performance on the classic linear regression task further validates the effectiveness of our framework.
	
	\item Different from gradient-based differentially private optimization algorithms, our SI optimization algorithms with DP exhibit some interesting phenomena under certain conditions. One is that their performance is not strictly affected by the privacy budget, and the other is that the performance is not necessarily worse than the non-private version.
	
	\item The whole study may provide researchers with a new perspective on DP, and promote the synergy between metaheuristic optimization community and privacy computing community.
\end{enumerate}

The rest of the paper is organized as follows. In Section~\ref{sec.preliminary}, we introduce some necessary preliminaries on DP and SI algorithms. In Section~\ref{sec.DPSIAF}, we propose the DPSIAF. The theoretical analyses are conducted in Section~\ref{sec.theo_ana}. The experimental results are reported and analyzed in Section~\ref{sec.experiments}, and we draw the conclusion in Section~\ref{sec.conclusion}.

\section{Preliminary}
\label{sec.preliminary}
In this section, we provide several preliminaries about DP and SI algorithms that will be involved in this work.

\subsection{Differential Privacy}
DP is a novel and promising privacy-preserving model \cite{Dwork2008Differential}. Different from classic privacy model like $k$-anonymity \cite{Sweeney2002K}, it ignores the background knowledge of the adversary and provides a provable guarantee. By introducing the perturbation, DP can ensure that any output related to the dataset is independent of whether a certain record is in the dataset. As a result, even if the adversary knows all but the target record, the information of that record still cannot be inferred. Some basic definitions and theorems about DP are formally described as follows \cite{Dwork2014Algorithmic}.
\begin{definition}[$\epsilon$-DP]
	An algorithm $\mathcal{M}$ satisfies $\epsilon$-DP, if and only if for any pair of neighboring datasets $D$ and $D'$ that differ on one record, and for all measurable subsets $\mathcal{S} \subseteq \mathcal{R}$, we have
	\begin{equation}
		Pr[\mathcal{M}(D) \in \mathcal{S}] \leq \exp(\epsilon) \cdot Pr[\mathcal{M}(D') \in \mathcal{S}], 
	\end{equation}	
	where $\epsilon \geq 0$ refers to the privacy budget, $\mathcal{R}$ denotes the set of all possible outputs and $Pr[\cdot]$ measures the probability of outputs.
\end{definition}

In above inequality, $\epsilon$ controls the privacy level of the algorithm. The smaller $\epsilon$ represents the higher privacy level, which means more perturbation needs to be introduced. To determine the magnitude of the perturbation, the sensitivity, a concept that can describe the maximal difference of a function may have on the neighboring datasets, is defined. 
\begin{definition}[Sensitivity]
	\label{eq.sensitivity}
	Given a pair of neighboring	datasets $D$ and $D'$, the sensitivity of a function $f:D \rightarrow \mathcal{R}$ is defined as
	\begin{equation}
		\Delta{f} = \max \limits_{D,D'} \left\|f(D) - f(D')\right\|_{1},
	\end{equation}
	where $\Delta{f}$ is the sensitivity and it is only related to the type of function $f(\cdot)$.
\end{definition}

To satisfy the definition of DP, two prevalent randomization mechanisms, i.e., Laplace mechanism and exponential mechanism, are used for numeric and non-numeric outputs, respectively. The former adds independent noises sampled from the Laplace distribution to the true answer, and the latter selects the final output probabilistically according to the quality of each possible output. Since this work adopts the exponential mechanism, we hereby only give its formal definition.
\begin{definition}[Exponential Mechanism]
	Let $q(D, r)$ be a score function that evaluates the quality of output $r \in \mathcal{R}$ over dataset D. The algorithm $\mathcal{M}$ satisfies $\epsilon$-DP if
	\begin{equation}
		\label{eq.EM}
		\mathcal{M} = \Big\{\text{return} \ r \ \text{with probability} \propto\ \frac{\epsilon q(D,r)}{2\Delta{q}} \Big\},
	\end{equation}
	where $\Delta{q}$ is the sensitivity of score function $q(D,r)$.
\end{definition}

In addition, a complex privacy-preserving problem generally requires multiple applications of DP to solve. In order to ensure that the privacy level of the whole process is controlled within the given privacy budget $\epsilon$, several properties of DP can be reasonably used to allocate the entire budget to each step. The involved properties in this work are given below.
\begin{theorem}[Sequential Composition]
	\label{Theorem.Sequential Composition}
	Suppose that a set of privacy algorithms $\mathcal{M}$=\{$\mathcal{M}_1$, $\cdots$, $\mathcal{M}_k$\} are sequentially performed on a dataset, and each $\mathcal{M}_{i}$ provides an $\epsilon_{i}$-DP, then $\mathcal{M}$ will provide $\sum_{i=1}^{k}\epsilon_{i}$-DP.
\end{theorem} 

This theorem offers a privacy guarantee for a sequence of differentially private computations, it indicates that the privacy budget for each step is additive when a series of differentially private operations are performed sequentially on a dataset.

\begin{theorem}[Post-Processing]
	\label{Theorem.Post-Processing}
	Given algorithm $\mathcal{M}_1$ that satisfies $\epsilon$-DP on the dataset $D$,  then for any algorithm $\mathcal{M}_2$, the composition of $\mathcal{M}_1$ and $\mathcal{M}_2$, i.e., $\mathcal{M}_2(\mathcal{M}_1(D))$, satisfies $\epsilon$-DP.
\end{theorem}  

This theorem states that once an algorithm satisfies $\epsilon$-DP, then no matter what additional processing is performed on the output of this algorithm, the combination of this algorithm and additional processing still satisfies $\epsilon$-DP.

\subsection{Swarm Intelligence Algorithm}
Over the last two decades, evolutionary computation have become popular for solving optimization problems in various fields due to their attractive features such as simplicity and flexibility. 
Technically, evolutionary computation involves metaheuristic optimization algorithms. 
When solving an optimization problem, different from other algorithms, metaheuristic optimization algorithms represent the positions of individuals in the population as feasible solutions, and iteratively update these positions according to some heuristic strategies to obtain the optimal solution.
Depending on different inspirations, metaheuristic optimization algorithms involved in the evolutionary computation can be roughly classified into three categories. that is, evolutionary theory based algorithms, physical phenomena based algorithms, and SI based algorithms.

Among these three categories of algorithms, as we mentioned in the introduction, this work chooses SI algorithms as the study object, because they have some unique properties when compared with the other two kinds of algorithms \cite{Tang2021Review}. 
For example, the individuals in the population preserve the information about the search space over the course of iteration, and they also share this information with each other. Meanwhile, they have fewer parameters to adjust and fewer operators to perform. These properties make SI based algorithms easier to understand and implement, and may perform better in the face of various optimization problems. 
Since there are numerous SI algorithms, we hereby only briefly describe the main ideas of the four classic and popular algorithms involved in this work.

Kennedy and Eberhart proposed the PSO algorithm \cite{Kennedy1995Particle} in 1995 by mimicking the foraging behavior of bird flock. In the algorithm, the birds in the flock aim to fly around in a search space to find food, i.e., optimal solution. Initially, the birds are randomly scattered in the space. Then, as the iteration progresses, each bird continuously updates itself by tracking two best positions that it and the flock have found so far, until the stopping condition is reached.

Inspired by the pack hunting of grey wolves, Mirjalili \textit{et al.} proposed the Grey Wolf Optimizer (GWO) in 2014 \cite{Mmirjalili2014Grey}. There is a strict social hierarchy within the gray wolves. When they hunt, the entire wolf pack will constantly to search, surround and attack the prey under the command of several leading wolves until the prey is captured. GWO mimics this behavior, and in the iteration, each search agent updates itself with reference to the top three best positions in the population.

In 2016, inspired by the predatory behavior of humpback whales, Mirjalili \textit{et al.} proposed a novel SI algorithm, which is called the Whale Optimization Algorithm (WOA) \cite{Mirjalili2016Whale}. In the optimization process, the algorithm mimics three different movement models of humpback whales during predation, namely random movement, spiral movement and best-position-targeted movement to update the search agent.

Seagull Optimization Algorithm (SOA) is another novel SI algorithm proposed by Dhiman \textit{et al.} in 2019 \cite{Dhiman2019Seagull}. In the algorithm, each search agent updates itself by mimicking the migration and attack behaviors of seagulls in nature, that is, moving towards the best position in the population and performing spiral movement.

\section{Differentially Private Swarm Intelligence Algorithm Framework}
\label{sec.DPSIAF}
\subsection{Inspiration}
This work aims to discuss the combination of DP and SI by designing a general algorithm framework. 
To this end, two key problems need to be addressed. The first is, how to develop a non-private algorithm into the differentially private version? The second is how to ensure the generality of the development scheme?

For the first problem, according to the current study of DP in deep learning and machine learning \cite{Li2021When}, there are about four ways to achieve DP in an algorithm, and they are input perturbation, output perturbation, objective perturbation, and gradient perturbation. As the name implies, the four ways perturb the original data, output results, objective function and gradient information, respectively. 
For a SI algorithm, one of the most important properties is the flexibility, which makes it independent of the optimization problems. We hope to keep this attractive and unique property for its differentially private version, especially in the context of optimization computation outsourcing. Thus, the gradient perturbation seems to be more appropriate than the first three ways. 
Noticeably, although the SI algorithms do not involve the concept of gradient, we can still refer to the idea of gradient perturbation to achieve DP by introducing the perturbation into the inner part of the algorithm.  

For the second problem, it is not easy to deal with, because different SI algorithms are inspired by different species, and this makes them follow different strategies in the optimization process. Thus, unlike the gradient-based optimization algorithms that are explicitly guided by gradients, it is difficult to condense a general strategy for a wide variety of SI algorithms.
\begin{figure}[]
	\centering
	\includegraphics[scale=0.5]{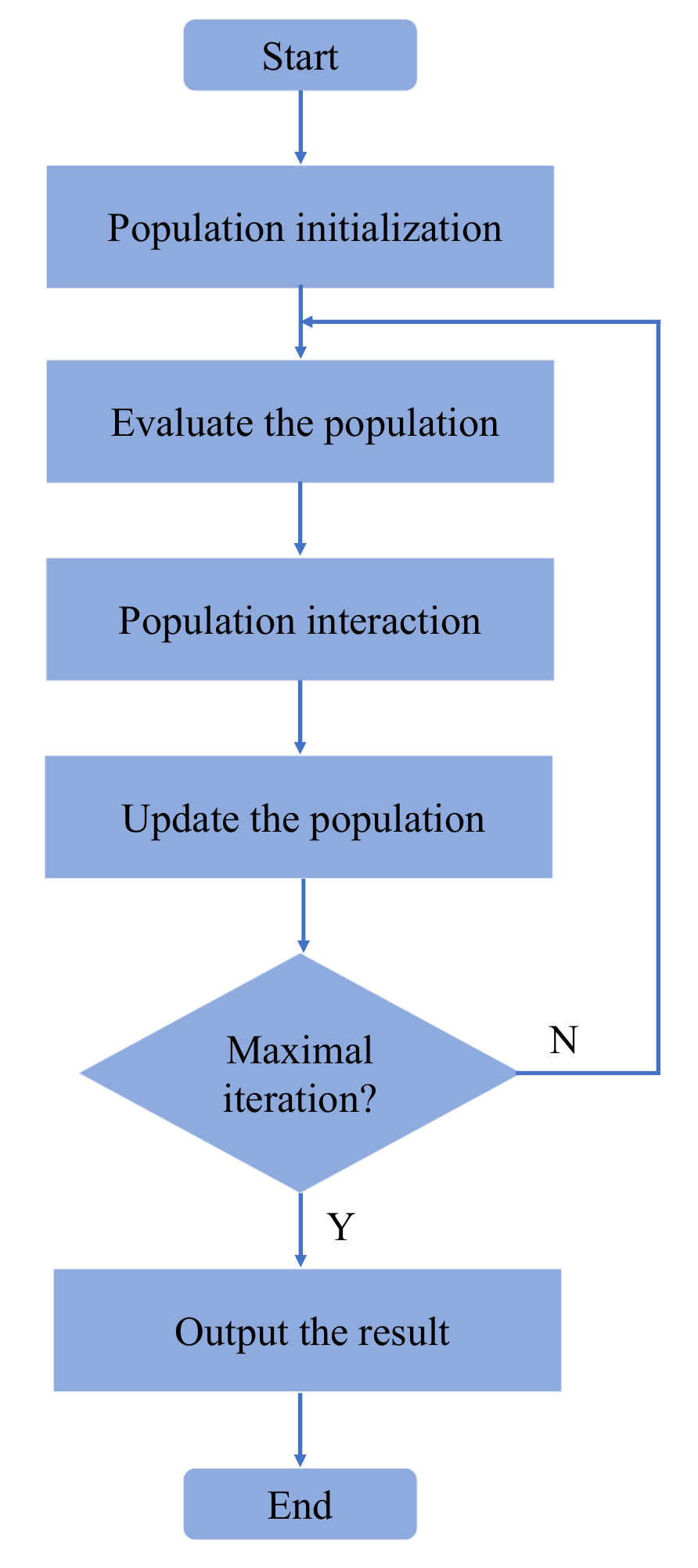}
	\caption{The general procedure of SI algorithms.}
	\label{fig.SIO_general procedure}
\end{figure}

To address the problem, we consider the concepts of generality and diversity from the perspective of the emergence of SI. According to our investigation on existing SI algorithms, actually, most of them follow a general procedure shown in Fig.~\ref{fig.SIO_general procedure}. At first, the population will be initialized. Then, the algorithm enters the iterative process. According to the objective function, each individual in the population will calculate the fitness, save and share it to update the global and/or personal best positions. After, the individuals will interact with each other and update their positions. The whole process will be repeated until the termination condition, usually the maximal iteration, is met. Finally, output the global best position that the algorithm can found. 

For different SI algorithms, the main difference between them is the behavior pattern. More specifically, the ways populations interact and update are different. Such differences lead to the emergence of different SI in different species, which is the source of diversity. 
In the context of DP, however, these parts generally do not access the original sensitive dataset, and they just use the outputs that produced and saved by previous parts. As a result, we only need to focus on the part that algorithms will access the sensitive dataset. The huge differences between different SI algorithms can be safely ignored, thus providing the basis for the generality of our designed framework.

\subsection{The Proposed Algorithm Framework}
\begin{algorithm}[t]
	\caption{DPSIAF}              
	\label{alg1}                        
	\begin{algorithmic}[1]
		\REQUIRE sensitive dataset $D$, objective function $f$, privacy budget $\epsilon$, maximal iteration $r$, and population size $m$.
		\ENSURE global best position $Gbest$.
		
		\STATE Outsourcer initializes the population $P$ with $m$ randomly generated vectors and sets necessary parameters and variables
		\STATE $\epsilon_{r}=\epsilon / r$
		\FOR{$j=1$ to $r$}
		\STATE Outsourcer sends the whole population $P$ to the user
		\STATE $Pbest = \mathcal{S}(f,D,P,m,\epsilon_{r})$	 \COMMENT{user updates the personal best positions $Pbest$ for each individual based on DP}
		\STATE User sends the whole $Pbest$ to the outsourcer
		\STATE Outsourcer updates the global best position $Gbest$ based on the fitness if there is a better position in the $Pbest$
		\FOR{$i=1$ to $m$}
		\STATE Outsourcer performs the population interaction operations 
		\STATE Outsourcer updates the position of individual $P[i]$ with refer to the $Gbest$ and/or the $Pbest[i]$
		\ENDFOR
		\ENDFOR
		\RETURN $Gbest$
	\end{algorithmic}
\end{algorithm}
\begin{algorithm}[t]
	\caption{DPSIAF: $\mathcal{S}$}              
	\label{alg2}                        
	\begin{algorithmic}[1]
		\REQUIRE objective function $f$, sensitive dataset $D$, population $P$, population size $m$, and privacy budget $\epsilon_r$.
		\ENSURE differentially private update personal best positions $Pbest$
		\STATE $\epsilon_{m}=\epsilon_{r} / m$
		\FOR{$i=1$ to $m$}
		\STATE User calculates fitness for $P[i]$ according to $D$ and $f$
		\STATE User selects a position between $P[i]$ and $Pbest[i]$ by applying the exponential mechanism with privacy budget $\epsilon_{m}$ to update $Pbest[i]$
		\ENDFOR
		\RETURN $Pbest$
	\end{algorithmic}
\end{algorithm}
Based on above inspirations, we propose the DPSIAF which is a general differentially private algorithm framework for SI algorithms. Any non-private SI algorithm is applicable for our framework, as long as it satisfies the following conditions.
\begin{enumerate}
	\item The algorithm accesses the sensitive dataset when it solves the optimization problem.
	\item The algorithm saves the global best position, personal best positions, and their corresponding fitness during the iteration.
	\item The algorithm updates the positions of population with refer to the global and/or personal best positions.
\end{enumerate}
These three constraint conditions, actually, are quite mild and most of the existing SI algorithms can easily meet them. In our assumed application scenario of optimization computation outsourcing, the pseudocodes of the proposed framework are shown in the Algorithms 1 and 2. The detailed steps are described as follows.
\begin{itemize} 
	\item Step 1: Outsourcer initializes $m$ individuals in the population $P$ with a random manner. Typically, $P$ is a matrix, and each row represents the position of an individual. Also, set the parameters and variables required by the algorithm including the privacy budget $\epsilon$ and maximal iteration $r$.
	\item Step 2: Outsourcer sends $P$ to the user and requests user's evaluation for each individual. Then, user calculates the fitness of each individual according to the sensitive dataset $D$ and the objective function $f$ that the user holds.   
	\item Step 3: For each individual, based on the exponential mechanism with $\epsilon_m$, user selects a position between its current position $P[i]$ and historical personal best position $Pbest[i]$ to update the $Pbest[i]$. Here, $Pbest$ is a matrix and it saves the personal best position that each individual can find so far, $i(i=1,2,\cdots,m)$ is the number of the individual $i$ in the $P$ and $Pbest$, and $\epsilon_{m}=\frac{\epsilon}{rm}$. Then, send $Pbest$ to the outsourcer.
	\item Step 4: Based on the fitness, outsourcer selects the best position from $Pbest$. Then, update the global best position $Gbest$ if the selected position is better than it. 
	\item Step 5: According to the behavior pattern of the SI algorithm, outsourcer performs the population interaction operations to obtain the information needed for the population updates.
	\item Step 6: Outsourcer updates the position of each individual with refer to the global and/or personal best positions.
	\item Step 7: Repeat Steps 2-6 until the maximum number of iterations is reached. Finally, output the $Gbest$ as the best solution that the algorithm can found in the context of DP.
\end{itemize}

According to the Algorithm 1, we can see that our framework still follows the general procedure of a non-private SI algorithm on the whole. However, the update of personal best position for each individual is more complicated because this process involves accessing the sensitive dataset. As a result, the perturbation needs to be introduced here to satisfy the definition of DP, and Algorithm 2 shows our operations for this purpose. 

Unlike the approach of the non-private algorithm, we do not perform the selection directly based on the fitness, but use the exponential mechanism. We divide the privacy budget $\epsilon_{r}$ into $m$ equal parts, and consume a part for each individual to select one position between the individual's current position and its historical personal best position under the DP. Accordingly, the better the fitness, the higher the probability of being selected. The final output position is taken as the individual's current personal best position, and ideally, it is the same as the non-private's output position. 

Furthermore, for the update of global best position, we keep the same with the non-private version and do not make any changes. Based on the fitness, we select a best position from all the current personal best positions and compare it with the historical global best position. If the selected position is better, then take it as the current global best position. Obviously, this operation is inherently tied to the personal best positions of all individuals. Thus, the output is also in the context of DP even though no additional operations are introduced here, because the personal best positions of all individuals, has already been perturbed before. This ensures certain SI algorithms that only involve the global best position in the population update are still privacy-preserving.

It is worth noting that referring to the latest progress in the field of DP, more advanced DP implementation mechanisms and privacy budget allocation strategies can undoubtedly be adopted here \cite{zhao2022survey}. However, since this work is the first attempt to combine SI and DP, it is more appropriate to choose the classic scheme. 
As an example, Algorithm 3 shows pseudocode of the differentially private particle swarm optimization algorithm (DPPSO) developed by our proposed framework.
\begin{algorithm}[]
	\caption{DPPSO algorithm} 
	\begin{algorithmic}[1]
		\REQUIRE sensitive dataset $D$, objective function $f$, privacy budget $\epsilon$, maximal iteration $r$, and population size $m$.
		\ENSURE global best position $Gbest$.
		\STATE Outsourcer randomly initializes the positions $P$ and velocities $V$ of particles in the population
		\STATE $c_1=c_2=2$
		\STATE $\epsilon_{m}=\epsilon/(rm)$
		\FOR{$j=1$ to $r$}
		\STATE Outsourcer sends $P$ to the user
		\FOR{$i=1$ to $m$}
		\STATE User calculates the fitness of $P[i]$ based on $D$ and $f$
		\STATE User updates $Pbest[i]$ by using the exponential mechanism with privacy budget $\epsilon_{m}$
		\ENDFOR
		\STATE User sends $Pbest$ to the outsourcer
		\STATE Outsourcer selects the best position from $Pbest$ according to the fitness, then updates $Gbest$ if the selected position is better than it
		
		\FOR{$i=1$ to $m$}
		\STATE $r_1=rand(0,1)$
		\STATE $r_2=rand(0,1)$
		\STATE $V[i] = V[i] + c_1r_1(Pbest[i] - P[i]) + c_2r_2(Gbest - P[i])$ \COMMENT{outsourcer updates the velocity of particle $i$}
		\STATE $P[i] = P[i] + V[i]$ \COMMENT{outsourcer updates the position of particle $i$}
		\ENDFOR
		
		\ENDFOR
		\RETURN $Gbest$
	\end{algorithmic}
\end{algorithm}

\section{Theoretical Analysis}
\label{sec.theo_ana}
We hereby perform the theoretical analyses to show the effectiveness and correctness of our proposed framework.  
\begin{theorem}
	\label{Theorem.Alg.2}
	The Algorithm 2 satisfies $\epsilon_{r}$-DP.
\end{theorem}

\begin{IEEEproof}
	For each individual, the Algorithm 2 consumes privacy budget $\epsilon_{m}$ to update its personal best positions in the way of exponential mechanism. Without loss of generality, we let $\mathcal{C}$ represents this mechanism, its input and output are dataset $D$ and $w \in O$ respectively, and the used score function is $q(D,w)$. According to \eqref{eq.EM}, we have
	\begin{align}
		&\frac{\Pr[\mathcal{C}(D)=w]}{\Pr[\mathcal{C}(D')=w]} \notag \\
		=&\frac{\exp\left(\frac{\epsilon_{m} q(D,w)}{2\Delta q}\right)}
		{\sum_{w'\in O}\exp\left(\frac{\epsilon_{m} q(D,w')}{2\Delta q}\right)}
		\Biggl/
		\frac{\exp\left(\frac{\epsilon_{m} q(D',w)}{2\Delta q}\right)}{\sum_{w'\in O}\exp\Big(\frac{\epsilon_{m} q(D',w')}{2\Delta q}\Big)} \notag \\
		=&\frac{\exp\left(\frac{\epsilon_{m} q(D,w)}{2\Delta q}\right)}
		{\sum_{w'\in O}\exp\left(\frac{\epsilon_{m} q(D,w')}{2\Delta q}\right)}\frac{\sum_{w'\in O}\exp\left(\frac{\epsilon_{m} q(D',w')}{2\Delta q}\right)}{\exp\left(\frac{\epsilon_{m} q(D',w)}{2\Delta q}\right)} \notag \\
		=&\left(\frac{\exp\left(\frac{\epsilon_{m} q(D,w)}{2\Delta q}\right)}{\exp\left(\frac{\epsilon_{m} q(D',w)}{2\Delta q}\right)}\right)\left(\frac{\sum_{w'\in O}\exp\left(\frac{\epsilon_{m} q(D',w')}{2\Delta q}\right)}{\sum_{w'\in O}\exp\left(\frac{\epsilon_{m} q(D,w')}{2\Delta q}\right)}\right) \notag \\
		=&\exp\left(\frac{\epsilon_{m}\left(q(D,w)-q(D',w)\right)}{2\Delta q}\right) \notag \\
		& \times  \left(\frac{\sum_{w'\in O}\exp\left(\frac{\epsilon_{m} q(D',w')}{2\Delta q}\right)}{\sum_{w'\in O}\exp\left(\frac{\epsilon_{m} q(D,w')}{2\Delta q}\right)}\right).
		\label{eq.DPPSO_1}
	\end{align}
	Then, based on \eqref{eq.sensitivity}, the sensitivity of the score function $q(\cdot,\cdot)$ can be defined as
	\begin{equation}
		\Delta q = \max \limits_{w \in O, D, D'} \left\|q(D,w)-q(D',w)\right\|_1.
	\end{equation}
	Thus, \eqref{eq.DPPSO_1} can be rewritten as
	\begin{align}
		&\frac{\Pr[\mathcal{C}(D)=w]}{\Pr[\mathcal{C}(D')=w]} \notag \\
		\leq &\exp\left(\frac{\epsilon_{m} \Delta q}{2 \Delta q}\right) \left(\frac{\sum_{w'\in O}\exp\left(\frac{\epsilon_{m} q(D',w')}{2\Delta q}\right)}{\sum_{w'\in O}\exp\left(\frac{\epsilon_{m} q(D,w')}{2\Delta q}\right)} \right) \notag\\
		\leq &\exp\left(\frac{\epsilon_{m}}{2}\right) \left( \frac{\sum_{w'\in O}\exp\left(\frac{\epsilon_{m} q(D',w')}{2\Delta q}\right)}{\sum_{w'\in O}\exp\left(\frac{\epsilon_{m} q(D,w')}{2\Delta q}\right)} \right).
		\label{eq.DPPSO_2}
	\end{align}
	
	Compare \eqref{eq.DPPSO_1} and \eqref{eq.DPPSO_2}, we can see that 
	\begin{align}
		\frac{\exp\left(\frac{\epsilon_{m} q(D,w)}{2\Delta q}\right)}{\exp\left(\frac{\epsilon_{m} q(D',w)}{2\Delta q}\right)} =&\exp\left(\frac{\epsilon_{m}(q(D,w)-q(D',w))}{2\Delta q}\right) \notag \\
		\leq& \exp\left(\frac{\epsilon_{m} \Delta q}{2 \Delta q}\right) \notag \\
		\leq& \exp\left(\dfrac{\epsilon_{m}}{2}\right).
	\end{align}
	According to the symmetry of neighboring dataset, we also have 
	\begin{align}
		\forall w', \exp\left(\frac{\epsilon_{m} q(D', w')}{2 \Delta q}\right) \leq \exp\left(\frac{\epsilon_{m}}{2}\right) \exp\left(\frac{\epsilon_{m} q(D, w')}{2\Delta q}\right).
	\end{align}
	Thus, \eqref{eq.DPPSO_2} can be rewritten as
	\begin{align}
		&\frac{\Pr[\mathcal{C}(D)=w]}{\Pr[\mathcal{C}(D')=w]} \notag \\
		\leq &\exp\left(\frac{\epsilon_{m}}{2}\right) \left( \frac{ \sum_{w'\in O} \exp\left(\frac{\epsilon_{m}}{2}\right) \exp\left(\frac{\epsilon_{m} q(D, w')}{2\Delta q}\right)}{\sum_{w'\in O}\exp\left(\frac{\epsilon_{m} q(D,w')}{2\Delta q}\right)} \right)\notag \\
		\leq &\exp\left(\frac{\epsilon_{m}}{2}\right) \exp\left(\frac{\epsilon_{m}}{2}\right) \left(\frac{ \sum_{w'\in O} \exp\left(\frac{\epsilon_{m} q(D, w')}{2\Delta q}\right)}{\sum_{w'\in O}\exp\left(\frac{\epsilon_{m} q(D,w')}{2\Delta q}\right)}\right) \notag \\
		\leq & \exp(\epsilon_{m}).
	\end{align}
	Therefore, the mechanism $\mathcal{C}$ satisfies $\epsilon_{m}$-differential privacy, which means it outputs $w \in O$ with probability proportional to $\exp\left(\frac{\epsilon_{m} q(D, w)}{2 \Delta q}\right)$. 
	
	Furthermore, since mechanism $\mathcal{C}$ is executed $m$ times by the Algorithm 2, and in terms of the composability property of differential privacy shown in Theorem \ref{Theorem.Sequential Composition}, the Algorithm 2 satisfies $(m
	\cdot\epsilon_{m})$-DP, i.e., $\epsilon_{r}$-DP. The proof is completed.
\end{IEEEproof}

In addition, based on Theorem \ref{Theorem.Alg.2}, the following theorem can be obtained.
\begin{theorem}
	The Algorithm 1 satisfies $\varepsilon$-DP.
\end{theorem}

\begin{IEEEproof}
	In Algorithm 1, the operations after Algorithm 2 are to determine the current global best position, perform the  interactions and update the position of each individual. Since these operations do not access the sensitive dataset again, there is no need to consume the privacy budget to perturb the primitive results. 
	
	Based on the post-processing property of DP, i.e., Theorem \ref{Theorem.Post-Processing}, it is clear that the single iteration of Algorithm 1 satisfies $\varepsilon_{r}$-DP. Then, similarly, based on the composability property, the whole iterations will consume $(r \cdot \epsilon_{r})$ privacy budget. Therefore, the Algorithm 1 satisfies $\varepsilon$-DP, and the proof is completed.
\end{IEEEproof}

The above theoretical analyses demonstrate that, the proposed algorithm framework satisfies the definition of DP, and this demonstrates the effectiveness and correctness of our development scheme. As a result, the algorithms developed based on our framework own the ability of privacy protection. 

\section{Experimental Evaluation}
\label{sec.experiments}
Similar to the development of DPPSO, we apply the proposed framework to another three classic SI algorithms, i.e., GWO, WOA and SOA, and obtain their differentially private versions. They are named Differentially Private Grey Wolf Optimizer (DPGWO), Differentially Private Whale Optimization Algorithm (DPWOA), and Differentially Private Seagull Optimization Algorithm (DPSOA), respectively. To evaluate the performance of these algorithms, this section conducts experiments on the linear regression task and analyze their performance.

\subsection{Optimization Task}
As we said in Section~\ref{sec.intro}, since this work is the first attempt to combine SI and DP, we hereby use the linear regression, one of the most typical optimization tasks, as an application example to evaluate the performance of discussed algorithms. 

For a dataset $D$ that contains $n$ tuples $t_1,t_2, \cdots, t_n$ and $d+1$ attributes $X_1, X_2, \cdots, X_d, Y$. The liner regression,  roughly speaking, aims to find a set of parameters that can enables us to predict the value $y_i$ on attribute $Y$ of any tuple $t_i$ as accurate as possible based on its values $x_i=(x_{i1},x_{i2}, \cdots, x_{id})$ on attributes  $X_1, X_2, \cdots, X_d$. Mathematically, the linear regression can be described as 
\begin{equation}
	w^{*} = \mathop{\arg\min}\limits_{w} f (D,w)=\mathop{\arg\min}\limits_{w}\sum_{i=1}^{n}  (w^{T} x_{i} - y_i)^2,
\end{equation}
where $f(\cdot,\cdot)$ is the objective function and $w^{*}$ is a vector of $d$ real numbers.

In the exponential mechanism, we will use a score function to evaluate the quality of a candidate output. Note
that the exponential mechanism does not make any assumptions on the score function, thus we can simply define it based on the objective function of linear regression, that is 
\begin{equation}
	q(D, w)=-f(D, w)=-\frac{1}{n} \sum_{i=1}^{n}\left(w^{T} x_{i}-y_{i}\right)^{2}.
\end{equation}

Furthermore,  the sensitivity of this score function $\Delta q$ can be derived as follows.
\begin{align}
	\label{Eq.sensitivity_1}
	\Delta q &=\max _{w \in O, D,D'}\left\|q(D, w)-q\left(D', w\right)\right\|_{1} \notag \\
	&=\max _{w \in O, D,D'}\left\|\max q(D, w)-\min q\left(D', w\right)\right\|_{1}.
\end{align}
Since
\begin{align}
	&\max q(D, w)=-\min (-q(D, w)), \notag \\
	&\min q\left(D', w\right)=-\max \left(-q\left(D', w\right)\right),
\end{align}
we can rewrite \eqref{Eq.sensitivity_1} as
\begin{align}
	\Delta q &=\max _{w \in O, D,D'}\left\|\max q(D, w)-\min q\left(D', w\right)\right\|_{1} \notag\\
	&=\max _{w \in O, D,D'}\left\|-\min (-q(D, w))+\max \left(-q\left(D', w\right)\right)\right\|_{1} \notag \\
	&=\max _{w \in O, D,D'}\left\|\max \left(-q\left(D', w\right)\right)-\min (-q(D, w))\right\|_{1} \notag \\
	&=\max _{w \in O, D,D'}\left\|\max f(D, w)-\min f\left(D', w\right)\right\|_{1}.
\end{align}
Because $f\left(D', w\right)$ is the mean squared error function, we have $\min f\left(D', w\right)=0$.
Thus,
\begin{align}
	\label{Eq.sensitivity_2}
	\Delta q &=\max _{w \in O, D,D'}\left\|\max f(D, w)-0\right\|_{1} \notag \\
	&=\max _{w \in O, D}\|f(D, w)\|_{1} \notag \\
	&=\max _{w \in O}\left\|\left(\frac{1}{n} \sum_{i=1}^{n}\left(w^{T}x_{i}-y_{i}\right)^{2}\right)\right\|_{1}.
\end{align}

Without of loss generality, we assume that $-a \leq x_{ik}, y_i \leq a $ for $k \in \{1,2,\cdots,d\}$ and $i \in \{1,2,\cdots,n\}$, and \eqref{Eq.sensitivity_2} can be further rewritten as

\begin{align}
	\Delta q &=\max _{w \in O}\left\|\left(\frac{1}{n} \sum_{i=1}^{n}\left(w^{T}x_{i}-y_{i}\right)^{2}\right)\right\|_{1}  \notag\\
	&\leq\max _{w \in O}\left\|\left(\frac{1}{n} \sum_{i=1}^{n}\left(\sum_{j=1}^{d}\left|a w_{j}\right|+a\right)^{2}\right)\right\|_{1}  \notag \\
	&\leq\max _{w \in O} \left\|\left(\frac{1}{n}\left( n \left(\sum_{j=1}^{d}\left|a w_{j}\right|+a\right)^{2}  \right)\right)\right\|_{1} \notag \\
	&\leq\max _{w \in O} \left\| \left(\sum_{j=1}^{d}\left|a w_{j}\right|+a\right)^{2} \right\|_1.
\end{align}

\subsection{Experimental Setup}
In the experiments, we use two datasets from UCI Machine Learning Repository\footnote{https://archive-beta.ics.uci.edu/}, and they are \textit{Combined Cycle Power Plant} dataset and \textit{Gas Turbine CO and NOx Emission} dataset, respectively. 
The first dataset contains 9568 records collected from a combined cycle power plant when the plant is set to work with full load, and it has 4 features including \textit{Exhaust Vacuum} and \textit{Relative Humidity} to predict the net hourly electrical energy output of the plant. The second dataset contains 36733 records sampled from a gas turbine located in Turkey, and it has 10 features including the \textit{Carbon Oxide} and \textit{Nitrogen Oxides} to predict the turbine energy yield. The attributes of both datasets are real numbers and there are no missing values. Moreover, we normalize the values of each attribute to the range [-1, 1] and Table~\ref{tab.dataset property} summarizes the properties of used datasets.
\begin{table}[]
	\centering
	\caption{The properties of used datasets in the experiments}
	\label{tab.dataset property}
	\begin{tabular}{|c|c|c|}
		\hline
		Dataset & \begin{tabular}[c]{@{}c@{}}Combined Cycle\\  Power Plant\end{tabular} & \begin{tabular}[c]{@{}c@{}}Gas Turbine CO\\ and NOx Emission\end{tabular} \\
		\hline \hline
		Number of Records                               & 9568 & 36733 \\ \hline
		Dimensionality                                  & 5    & 11    \\ \hline
		\multicolumn{1}{|l|}{Attribute Characteristics} & Real & Real  \\ \hline
		Missing Values                                  & N/A  & N/A   \\ \hline
	\end{tabular}
\end{table}

The metric we used is the classical Root Mean Square Error (RMSE), and it is formulated as
\begin{equation}
	\text{RMSE} = \sqrt{\frac{\sum_{i=1}^{n}\left(\hat{y_{i}} - y_{i}\right)^2}{n}},
\end{equation}
where $\hat{y_{i}}$ and $y_{i}$ are the predicted value and actual value respectively, and $n$ is the number of records in the test set. Obviously, the lower the RMSE value, the higher the prediction accuracy.

The primitive algorithms and their parameter configurations are PSO ($c_1=c_2=2$), GWO ($a=2$), WOA ($b=1$), and SOA ($fc=2$). The parameter settings are determined by their respective original works, and naturally, their private versions, i.e., DPPSO, DPGWO, DPWOA, and DPSOA, also use the same parameter settings.
For all the algorithms, without loss of generality, the population size and the maximum number of iteration are respectively set to $m=100$ and $r=100$ to ensure that they have sufficient population size and can converge stably. Each experiment is performed 10-fold cross-validation 10 times and we report the average result. 
Last but not least, the seeds of the pseudo random number generator used in algorithms are randomly generated and then fixed. The purpose is to avoid the performance differences between private version and primitive version of the same algorithm are not caused by the different pseudo random sequences.

\subsection{Performance Analysis}
\begin{figure*}[]
	\flushleft 
	\subfloat[]{
		\includegraphics[width=0.25\textwidth]{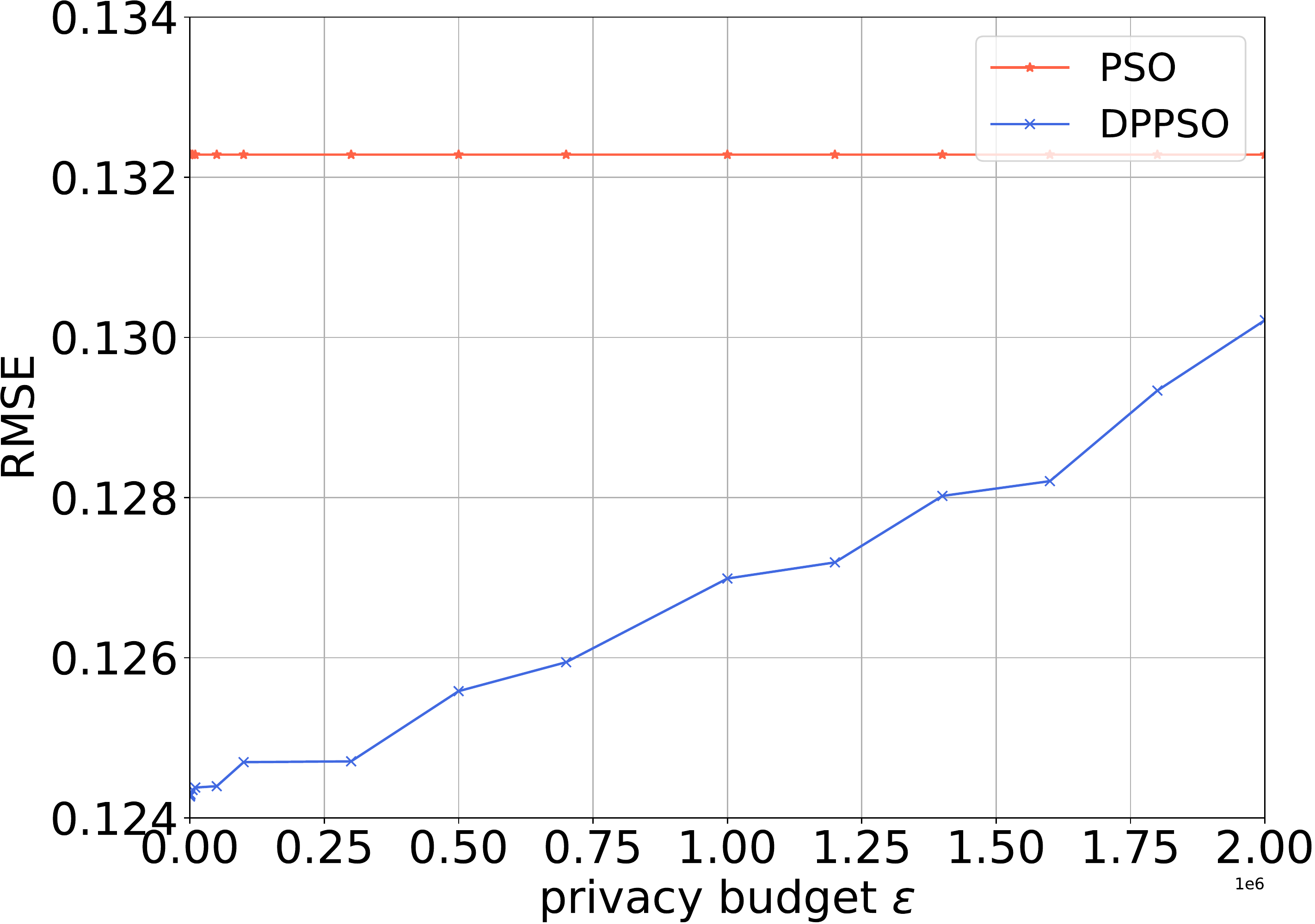}
		\label{fig.pso_d1}\hspace{-3.5mm}
	}
	\subfloat[]{
		\includegraphics[width=0.25\textwidth]{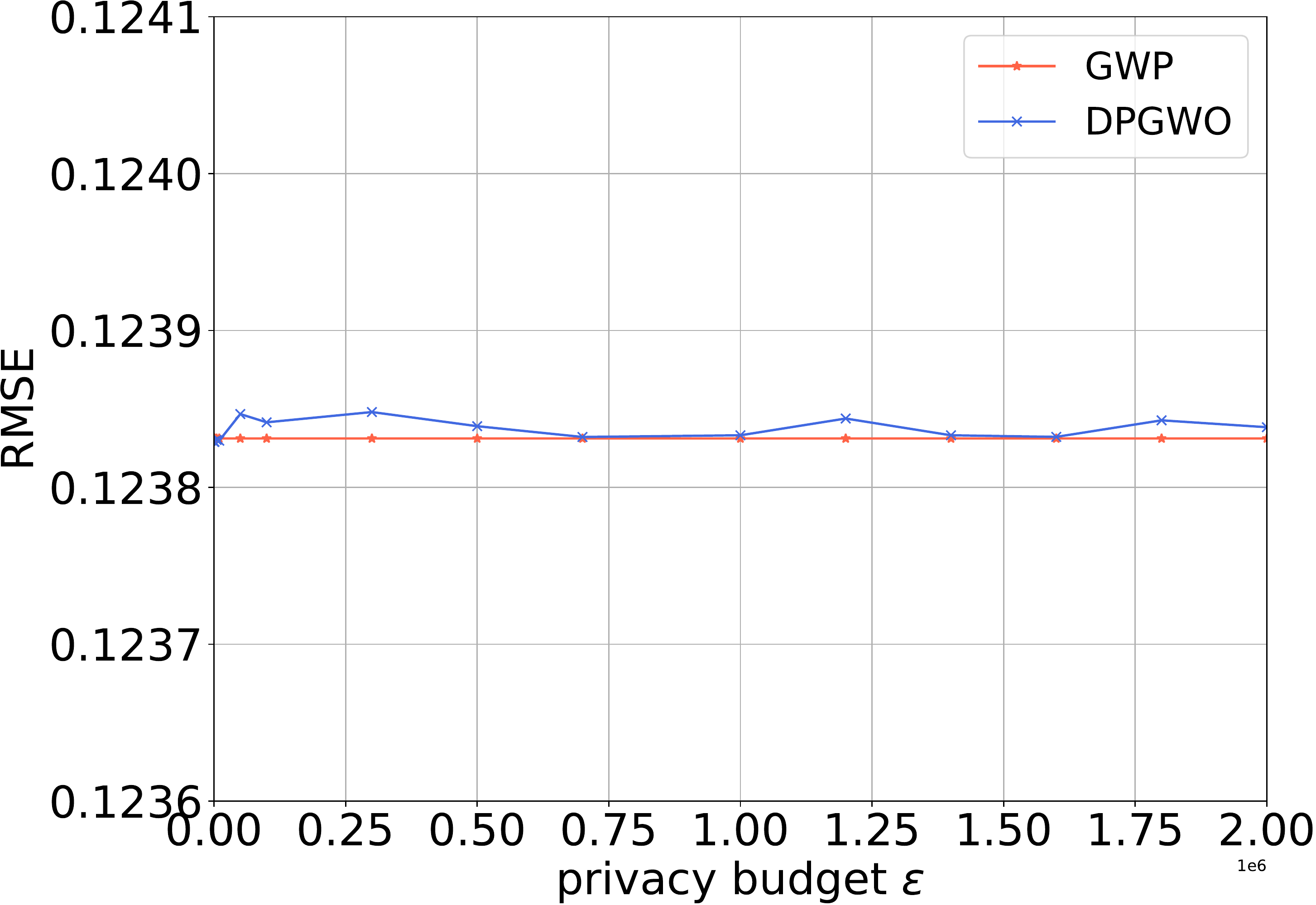}
		\label{fig.gwo_d1}\hspace{-3.5mm}
	}
	\subfloat[]{
		\includegraphics[width=0.25\textwidth]{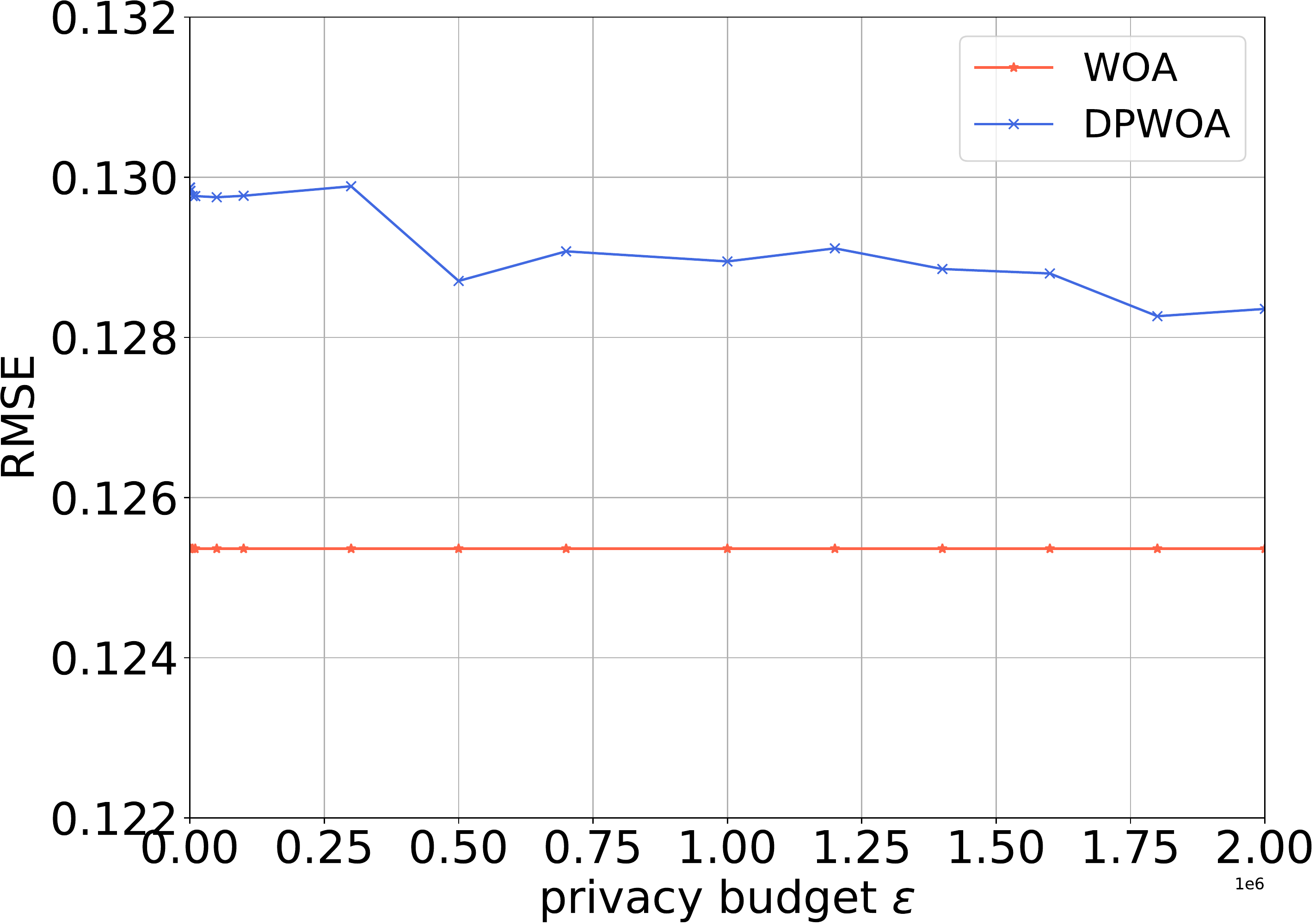}
		\label{fig.woa_d1}\hspace{-3.5mm}
	}
	\subfloat[]{
		\includegraphics[width=0.25\textwidth]{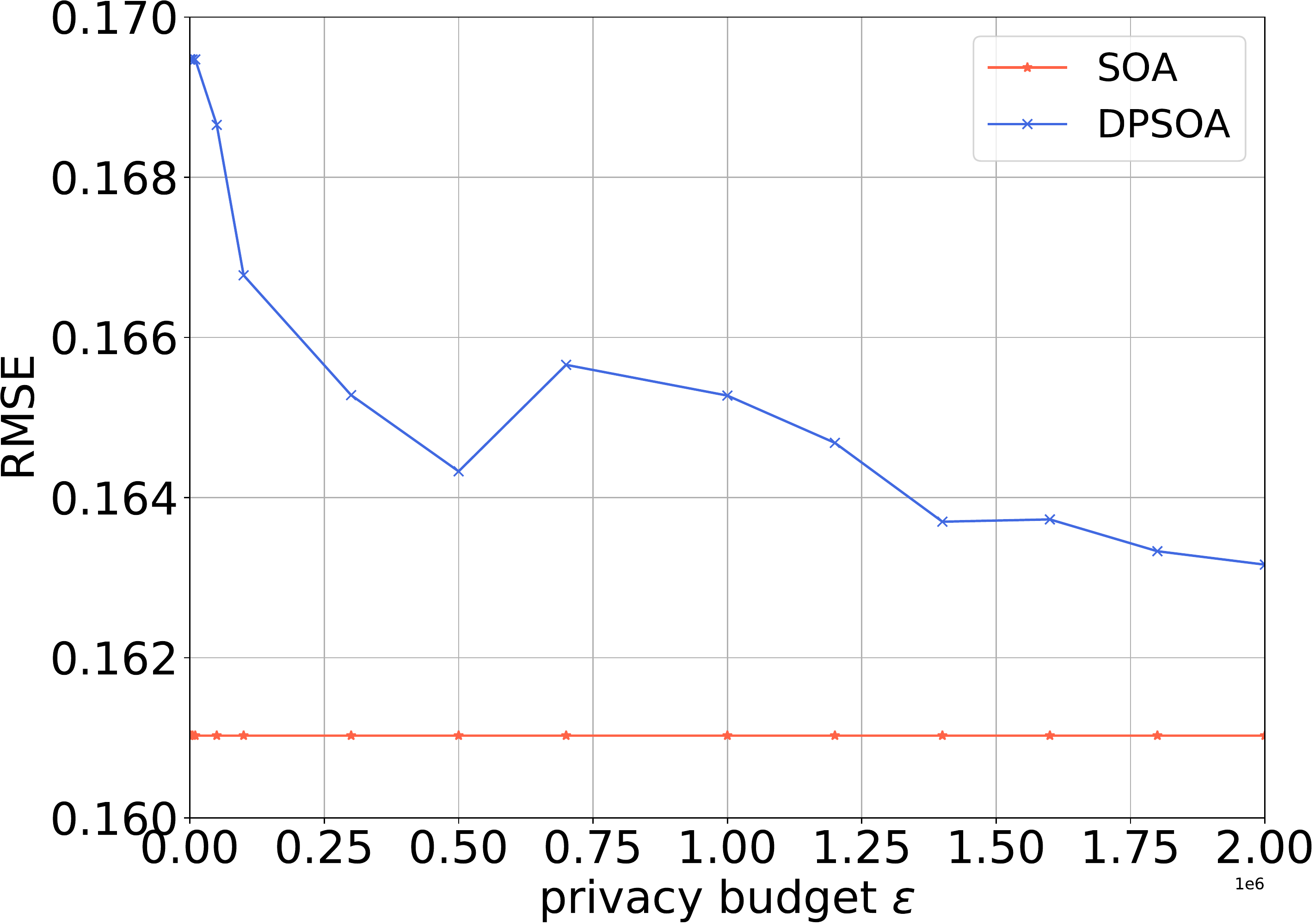}
		\label{fig.soa_d1}\hspace{-3.5mm}
	}
	\\
	\subfloat[]{
		\includegraphics[width=0.25\textwidth]{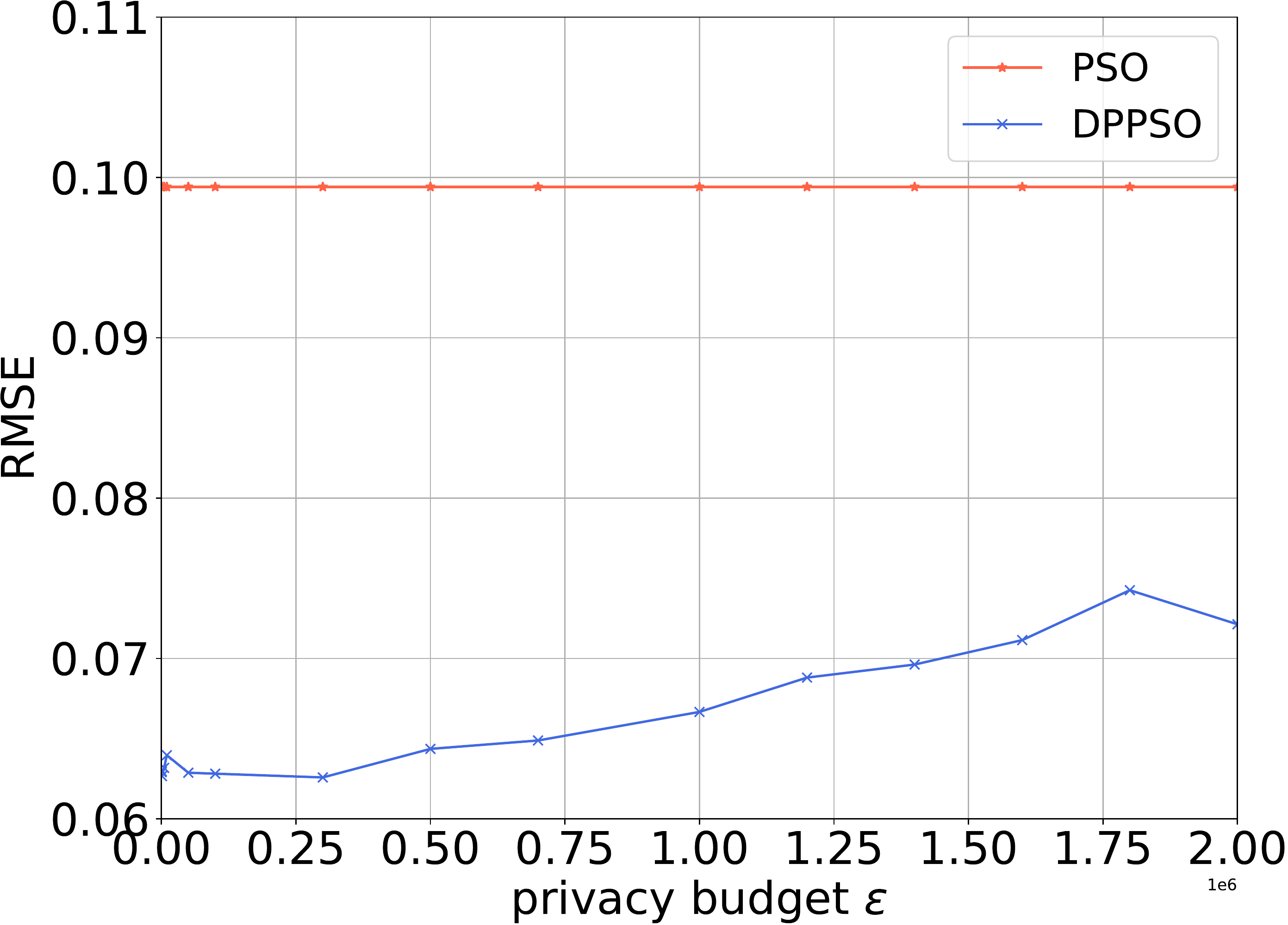}
		\label{fig.pso_d2}\hspace{-3.5mm}
	}
	\subfloat[]{
		\includegraphics[width=0.25\textwidth]{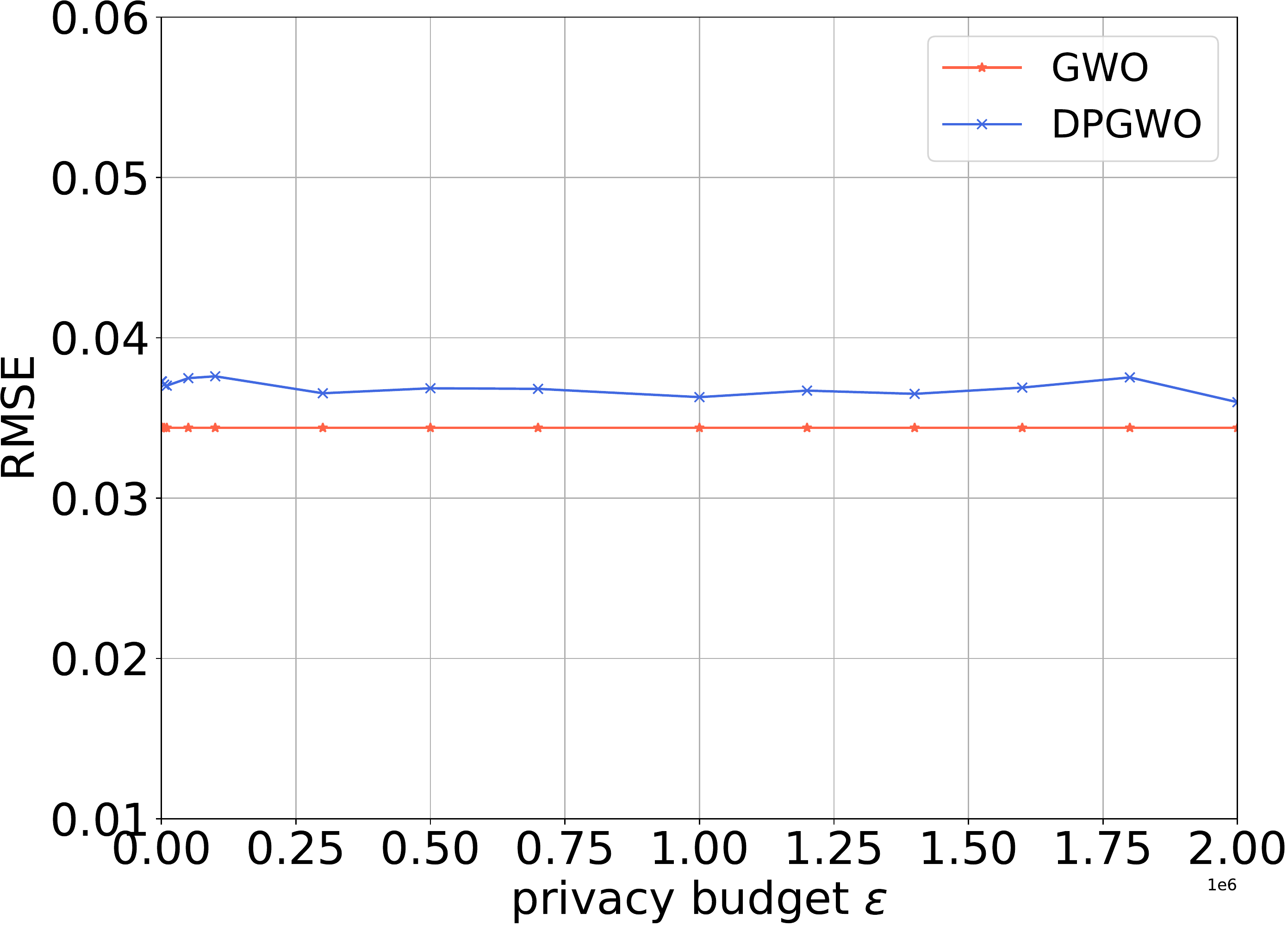}
		\label{fig.gwo_d2}\hspace{-3.5mm}
	}
	\subfloat[]{
		\includegraphics[width=0.25\textwidth]{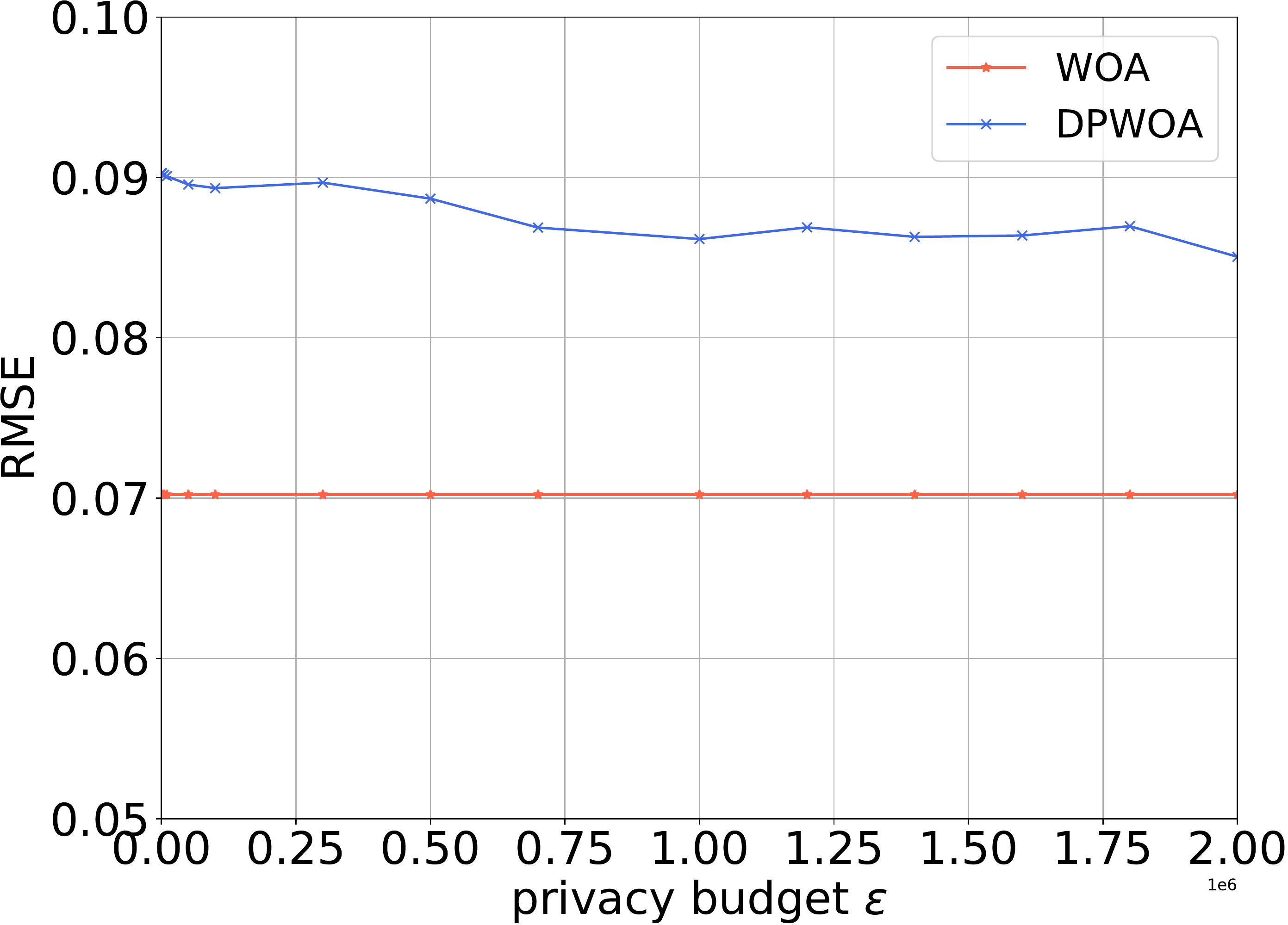}
		\label{fig.woa_d2}\hspace{-3.5mm}
	}
	\subfloat[]{
		\includegraphics[width=0.25\textwidth]{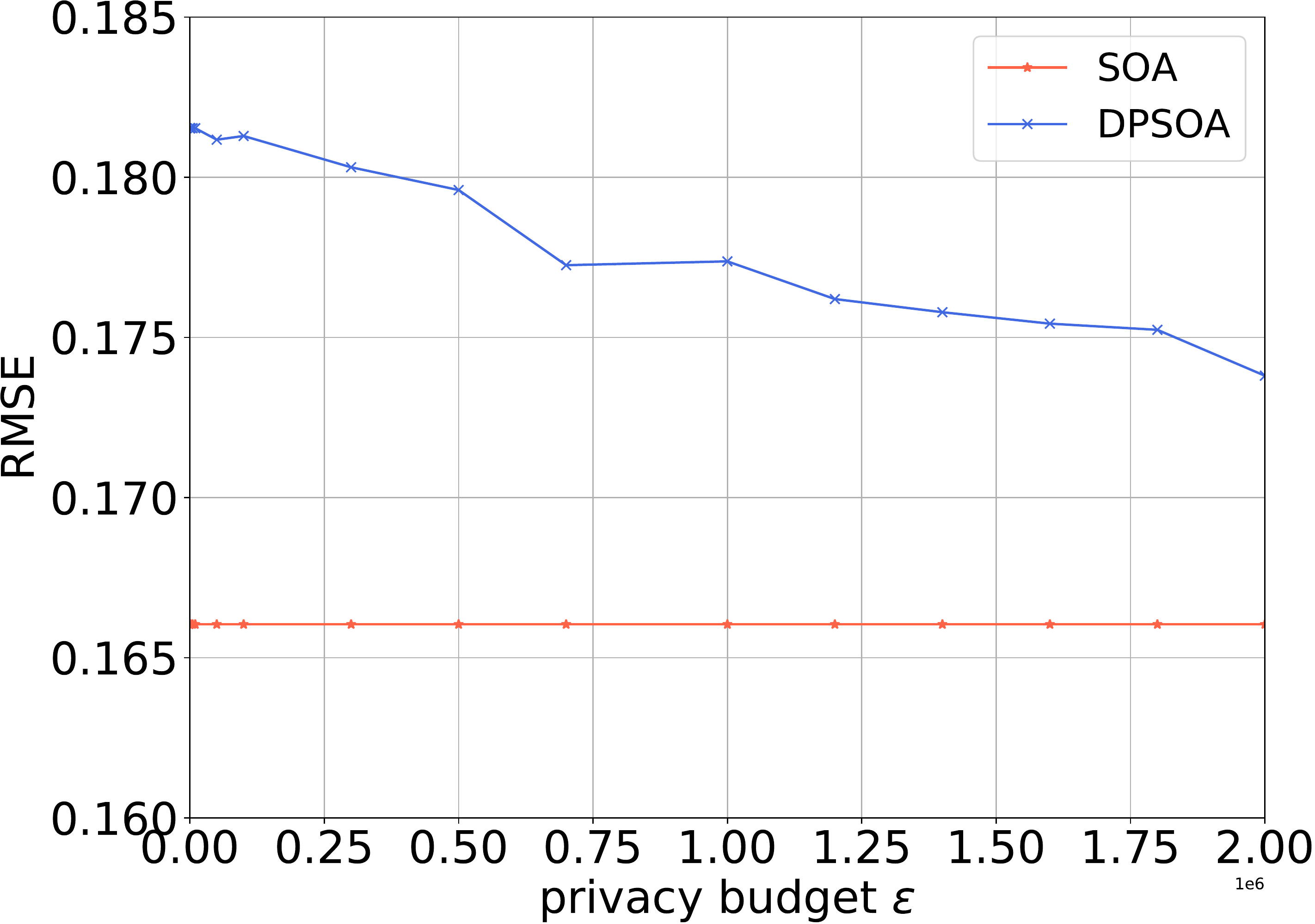}
		\label{fig.soa_d2}\hspace{-3.5mm}
	}
	\caption{The prediction accuracy of the discussed algorithms: The first row shows the results on the first dataset and the second row shows the results on the second dataset.}
	\label{fig.RMSE results}
\end{figure*}
Fig.~\ref{fig.RMSE results} plots the RMSE results of each algorithm as a function of the privacy budget $\epsilon$. As shown in the figure, the prediction accuracy of primitive algorithms remain unchanged for all values of $\epsilon$ since they do not involve the DP. 
For the private algorithms, interestingly, we observe some unusual phenomena. For DPWOA and DPSOA, their prediction accuracy is worse than their primitive counterparts, and the performance, overall, gets better as the privacy budget increases. For DPGWO, its prediction accuracy is slightly worse than its non-private version, and the performance does not change much as $\epsilon$ increases. Regarding DPPSO, it greatly outperforms the PSO, and on the whole, its performance becomes worse as the increase of the privacy budget. 
Moreover, we can see that the polylines of these private algorithms are not monotonic but have some undulations. Under some settings, the increase in the privacy budget does not necessarily lead to the improvement or degradation of the performance. 

Above experimental results are quite different from the conventional cognition. According to many reported studies, more specifically, the studies on integration of DP and gradient-based algorithms such as the DP-SGD \cite{Abadi2016Deep}, it is known that the introduction of DP will lead to the performance degradation of optimization algorithms, and the smaller the privacy budget, the more the degradation.
However, these conclusions do not seem to be fully applicable to our developed differentially private SI algorithms. The root cause, we argue, is that the working mechanisms of these two types of algorithms are essentially different. 

For gradient-based algorithms, they perform the search for solutions along the direction (or the opposite direction) of gradient in the process of optimization, because this is the direction in which the objective function changes the fastest. As we all know that, the gradient is deterministic. Once it is perturbed by DP, there is no doubt that the search direction of the algorithm will be biased. Naturally, the performance of gradient-based algorithms is worse than their non-private versions. And, smaller privacy budget means bigger perturbation, accordingly, their performance will degrade as the privacy budget decreases.
In stark contrast, the SI algorithms search for solutions by imitating the social behaviors of animals. Essentially, this approach is a heuristic strategy. Thus, the effect of DP on SI algorithms is not necessarily, exactly the same as that of the former. 

As the experimental results show, the performance of our private algorithms is not strictly affected by the size of the privacy budget. Locally, more perturbation may result in better performance and vice versa. 
Furthermore, since different SI algorithms have different behavior patterns, the DP may affect them in different degrees and natures. 
As we know, both SOA and WOA rely heavily on the global best location when updating the population. This said, the global best individual will strongly attract other individuals to its location. To some extend, these two algorithms are similar to the gradient-based algorithms. Accordingly, when the global best position is perturbed by DP, the performance of their private versions, i.e., DPSOA and DPWOA, will be negatively affected. 
For GWO, although it also relies heavily on the global best position, as we introduced in the Section II, this algorithm mimics the collective leadership system of the gray wolf pack, and the top three best individuals jointly guide the population. Thus, this algorithm is more robust. Naturally, the DPGWO suffers from the perturbation of DP to a much smaller extend, and it is relatively insensitive to changes in the privacy budget.

As for the PSO, Fig.~\ref{fig.RMSE results}(a) and (e) show that this algorithm is positively affected by DP. Different from above three algorithms, it simultaneously uses the personal best position and global best position to guide the update of the population. Considering that DPPSO has a more special performance and more complex behavior pattern, in the next subsection, we will design ablation experiments to further analyze it.

\subsection{Ablation Experiments}
\begin{figure}[]
	\centering
	\subfloat[]{
		\includegraphics[width=0.24\textwidth]{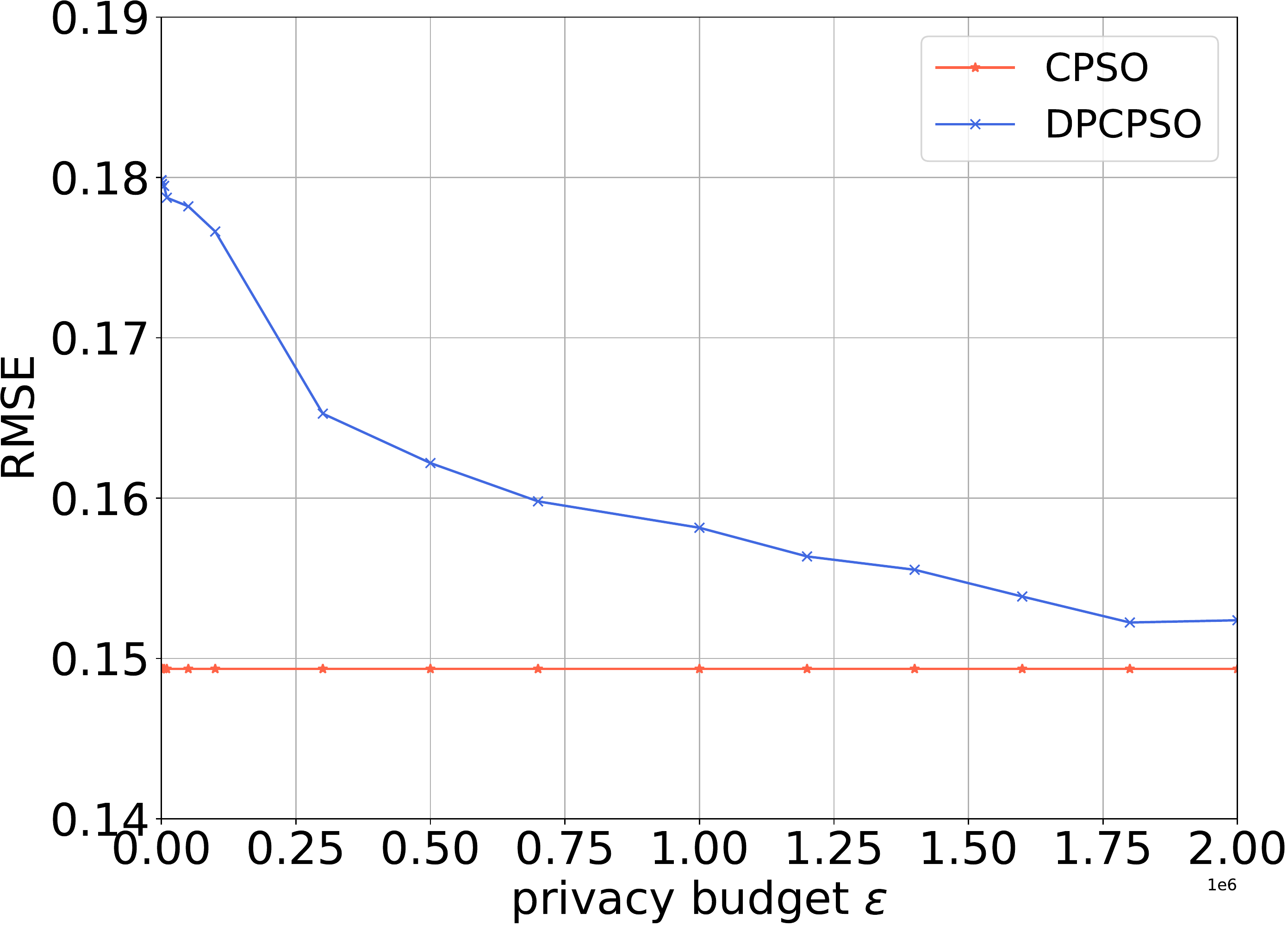}
		\label{fig.cpso_d1}\hspace{-3mm}
	}
	\subfloat[]{
		\includegraphics[width=0.24\textwidth]{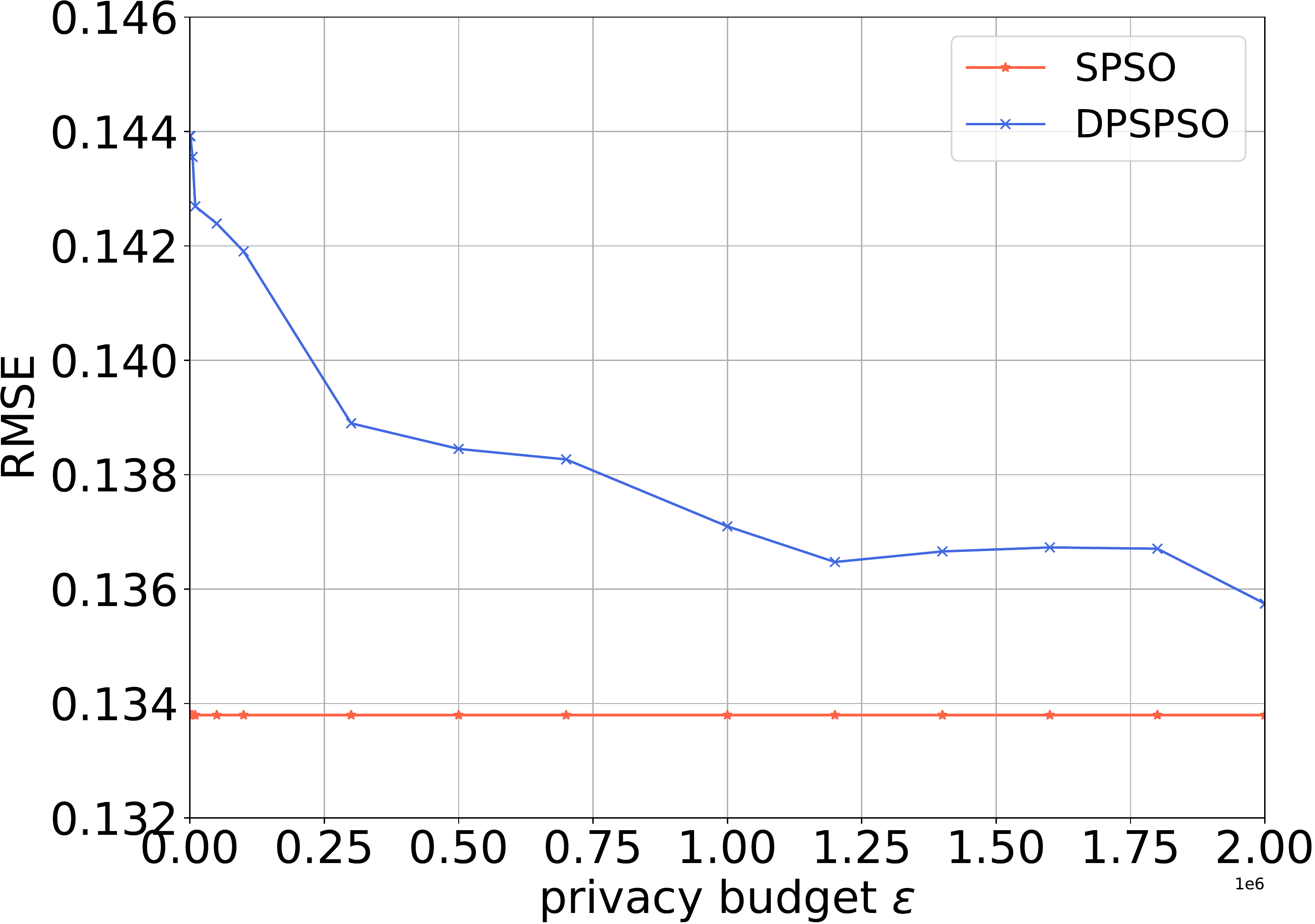}
		\label{fig.spso_d1}\hspace{-3mm}
	}\\
	\subfloat[]{
		\includegraphics[width=0.24\textwidth]{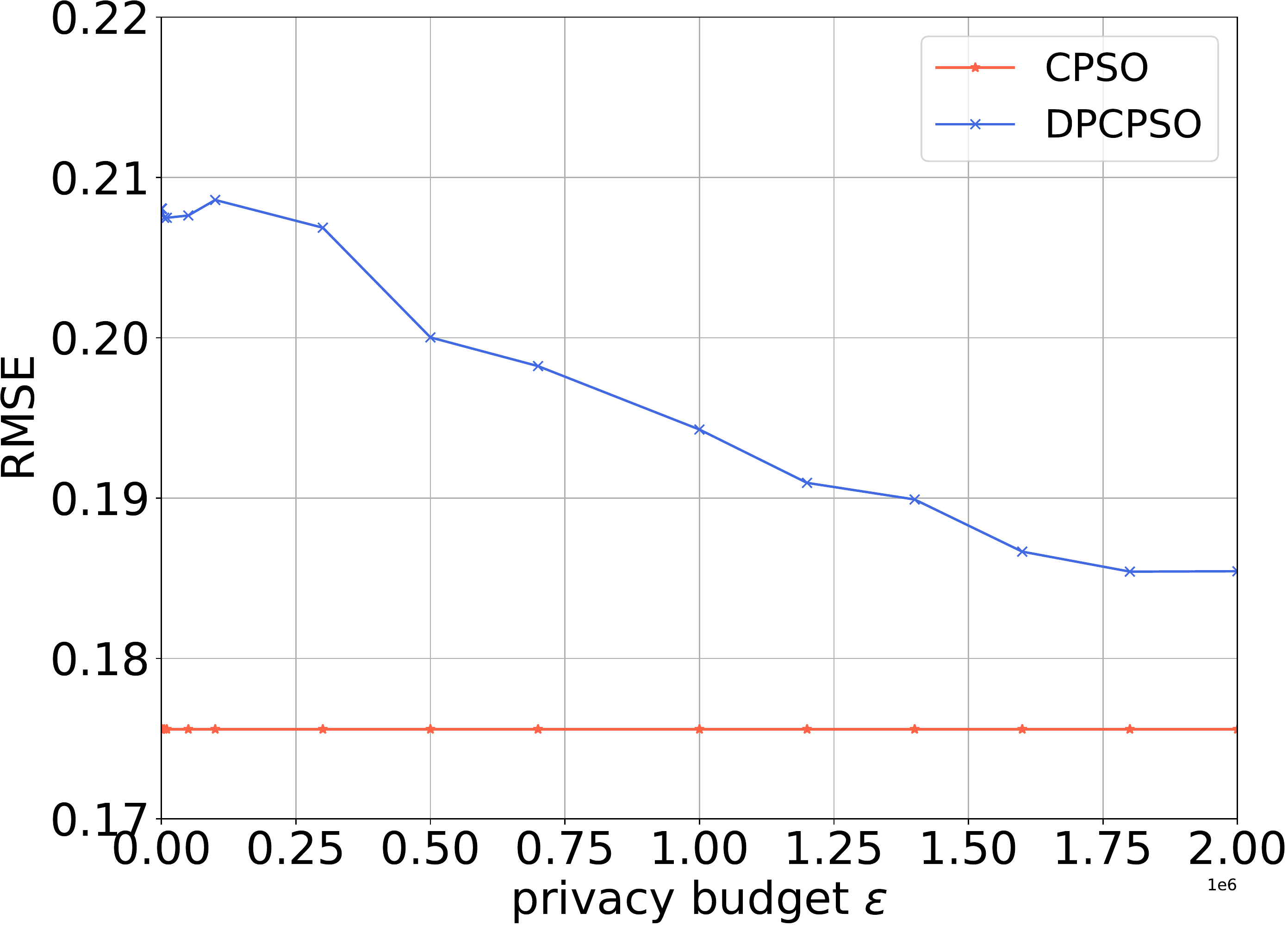}
		\label{fig.cpso_d2}\hspace{-3mm}
	}
	\subfloat[]{
		\includegraphics[width=0.24\textwidth]{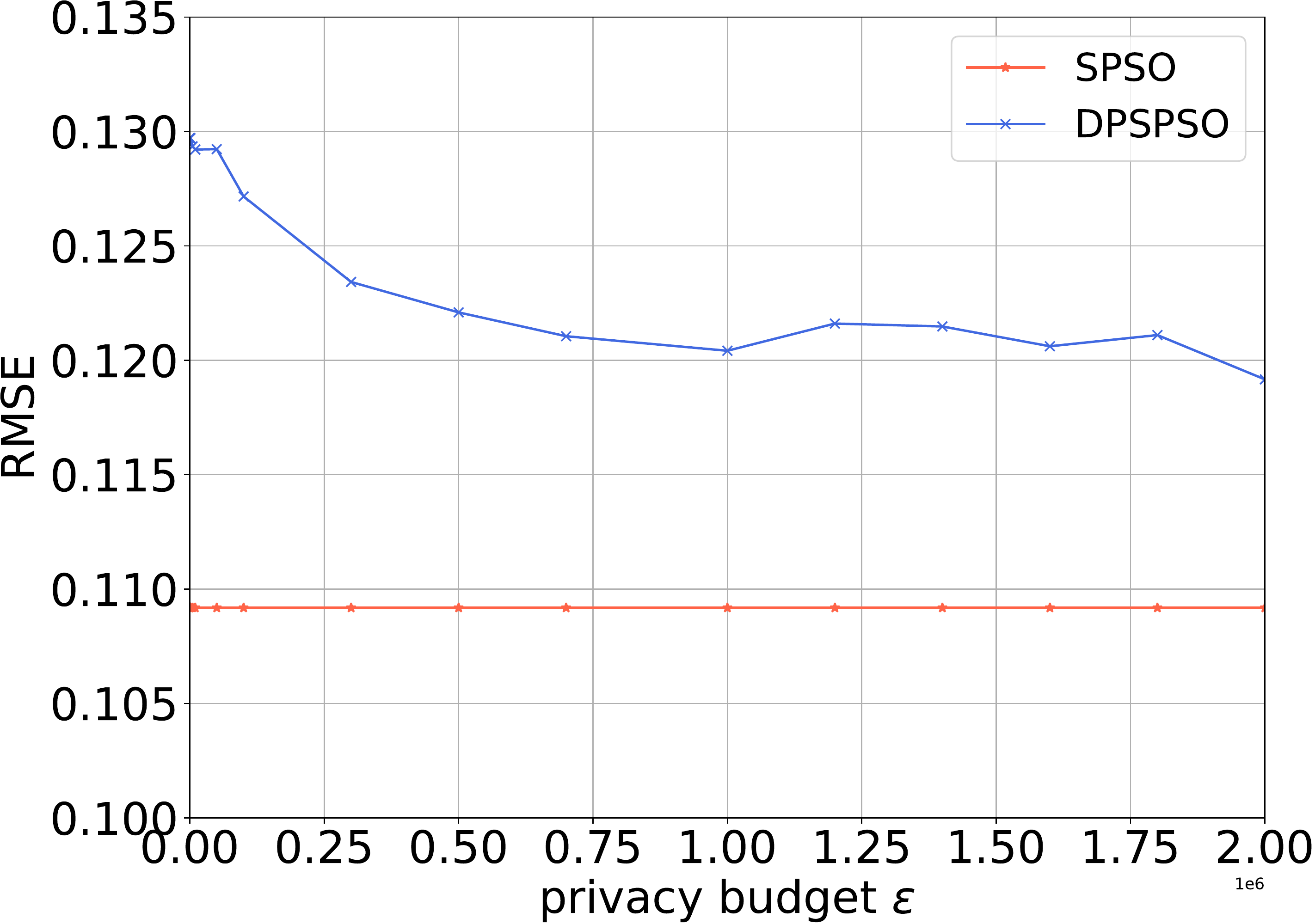}
		\label{fig.spso_d2}\hspace{-3mm}
	}
	\caption{The results of the ablation experiments: The first two subfigures show the results on the first dataset, and the other two shows the results on the second dataset.}
	\label{fig.ablation experiments}
\end{figure}

According to Algorithm 3, it is clear that both PSO and DPPSO update each individual through the following core mathematical expression, i.e., 
\begin{equation}
	\label{Eq.speed update of PSO}
	V[i] = V[i] + c_1r_1(Pbest[i] - P[i]) + c_2r_2(Gbest - P[i]).
\end{equation}
As we can see, it consists of three parts. Among them, the second part is the cognition part, which represents the private thinking of the individual itself, and the third part is the social part, which represents the collaboration among individuals \cite{Kennedy1997Particle}.

We hereby design ablation experiments to further compare and analyze the performance of PSOs with different behavior patterns after being perturbed by our framework. Specifically speaking, we modify \eqref{Eq.speed update of PSO} and let the PSO keep either the cognition part or the social part. The PSO with only cognition part is called Cognition Particle Swarm Optimization (CPSO), and accordingly, its private version is called Differentially Private Cognition Particle Swarm optimization (DPCPSO). Both of them use following expression to update the population.
\begin{equation}
	\label{Eq.speed update of CPSO}
	V[i] = V[i] + c_1r_1(Pbest[i] - P[i]).
\end{equation}
Similarly, the PSO with only social part is called Social Particle Swarm optimization (SPSO), and its private version is called Differentially Private Social Particle Swarm Optimization (DPSPSO). They use \eqref{Eq.speed update of SPSO} to update the population.
\begin{equation}
	\label{Eq.speed update of SPSO}
	V[i] = V[i] + c_2r_2(Gbest - P[i]).
\end{equation}
Other settings keep unchanged and the ablation experimental results are shown in Fig.~\ref{fig.ablation experiments}.

As one can see, the polylines of DPCPSO and DPSPSO are also not monotonic, which is consistent with that of the DPPSO. However, on the whole, the performance of these two algorithms is opposite to that of the DPPSO. First, their prediction accuracy is worse than their respective non-private versions, but the DPPSO is better than the PSO. Second, with the privacy budget increases, their performance goes up while the DPPSO comes down. 

The experimental results indicate that the DP has a positive effect on PSO only when its behavior pattern owns both cognition part and social part. Actually, for the PSO, the perturbation of DP is more like a special stimulus. This stimulus increases the diversity of the population, thereby enabling individuals to have a stronger ability to explore and exploit more promising regions in the search space. Meanwhile, the behavior pattern of PSO also makes the algorithm itself give positive feedback to this stimulus \cite{Wang2013Diversity}. Thus, to some extend, we argue that it is the difference in the capacity of the system to collectively respond to external perturbations that allows the PSO can be positively affected by the DP, while the CPSO and SPSO cannot \cite{Cavagna2022Marginal}. 

In addition, it is worth noting that the design of PSO itself involves some concepts of social psychology. From this perspective, when facing the challenge, groups that refer to both individual experience and leadership experience tend to perform better than groups that refer only to individual or leadership experience \cite{Kennedy1997Particle}. For them, the challenge is also an opportunity to learn and evolve \cite{Little2000Millennial}. Here, the perturbation of DP can be regarded as a kind of challenge because it brings the uncertainty to the group. 

\subsection{Discussion}
Above experimental results demonstrate that, the DP has different degrees and natures of the effects on SI algorithms with different behavior patterns. 
We note emphatically that, these results are obtained based on our developed algorithm framework and the linear regression task, different development schemes and/or optimization tasks may lead to different results. 
Nevertheless, since the proposed DPSIAF indeed satisfies the definition of DP, our study still reveals that under certain conditions, the use of DP does not necessarily lead to a decrease in the algorithm performance, but can actually improve it, although the precise definition of these conditions requires further in-depth research.

Undoubtedly, this provides a new perspective for the study of DP. For researchers in privacy computing, the privacy and the utility are not absolutely antagonistic. It is possible to design a private algorithm with tiny or even no loss in utility, at least for some specific problems. 
For researchers in metaheuristic optimization, DP has the potential to be a promising optimization tool like chaotic search \cite{Gao2021Chaotic} and Cauchy mutation \cite{Xu2019Enhanced}. The privacy budget is not limited to assessing the privacy level, nor does its value need to be restricted to a small range due to the data availability considerations.  And its allocation strategy may be used to tune the exploration phase and the exploitation phase in the optimization process.

In a nutshell, DP is rapidly gaining widespread attention of researchers in various fields. In addition to protecting the privacy, as a promising mathematical model, researchers also try to combine DP with other studies to expand its application fields and explore more uses. 
Some studies have already shown that the DP can effectively alleviate the overfitting in machine learning \cite{Dwork2015Reusable}, improve the stability in deep learning \cite{Zhao2020Privacy}, and avoid the malicious agent in game theory \cite{Lykouris2016Learning}. 
For this study, we are the first to focus on the integration of DP and SI, and some interesting phenomena shown in our experiments are also observed for the first time. This indicates that the DP can do more than the privacy. Although more and deeper works are needed in the future, we believe that this simple try may promote the synergy between different communities.

\section{Conclusion}
\label{sec.conclusion}
This paper proposed the DPSIAF for the first time, which aims to discuss the combination of DP and SI. 
By extracting the typical skeleton of SI algorithms, we leverage the exponential mechanism to perturb the selection of personal best position, and based on our framework, most existing SI algorithms can be easily developed into the corresponding private versions.
Theoretical analyses confirm the developed private algorithms satisfy the definition of $\epsilon$-DP. 
Furthermore, we take the classical linear regression as an example task and evaluate the performance of our developed four differentially private SI algorithms. 
The obtained results indicated that the performance of our algorithms is not strictly affected by the privacy budget, and more specially, the private PSO algorithm owns better performance than its non-private version.
From the perspectives of working mechanism, we empirically explained these interesting results and demonstrated that, DP may have different degrees and natures of the effects on SI algorithms with different behavior patterns, and it has the potential to serve as an optimization tool for some SI algorithms.

\bibliographystyle{IEEEtran}
\bibliography{my-TIFS.bib}

\begin{thebibliography}{10}
\providecommand{\url}[1]{#1}
\csname url@samestyle\endcsname
\providecommand{\newblock}{\relax}
\providecommand{\bibinfo}[2]{#2}
\providecommand{\BIBentrySTDinterwordspacing}{\spaceskip=0pt\relax}
\providecommand{\BIBentryALTinterwordstretchfactor}{4}
\providecommand{\BIBentryALTinterwordspacing}{\spaceskip=\fontdimen2\font plus
\BIBentryALTinterwordstretchfactor\fontdimen3\font minus
  \fontdimen4\font\relax}
\providecommand{\BIBforeignlanguage}[2]{{%
\expandafter\ifx\csname l@#1\endcsname\relax
\typeout{** WARNING: IEEEtran.bst: No hyphenation pattern has been}%
\typeout{** loaded for the language `#1'. Using the pattern for}%
\typeout{** the default language instead.}%
\else
\language=\csname l@#1\endcsname
\fi
#2}}
\providecommand{\BIBdecl}{\relax}
\BIBdecl

\bibitem{Cho2021Relational}
M.~Cho, T.~Kim, I.-J. Kim, K.~Lee, and S.~Lee, ``Relational deep feature
  learning for heterogeneous face recognition,'' \emph{IEEE Trans. Inf.
  Forensic Secur.}, vol.~16, pp. 376--388, 2021.

\bibitem{Jorge2022Live}
J.~Jorge, A.~Giménez, J.~A. Silvestre-Cerdà, J.~Civera, A.~Sanchis, and
  A.~Juan, ``Ieee-acm trans. audio speech lang.'' \emph{IEEE/ACM Transactions
  on Audio, Speech, and Language Processing}, vol.~30, pp. 148--161, 2022.

\bibitem{wang2022Hellinger}
Y.~Wang, X.~Zhao, Z.~Zhang, and L.~Y. Zhang, ``A collaborative filtering
  algorithm based on item labels and hellinger distance for sparse data,''
  \emph{Journal of Information Science}, vol.~48, no.~6, pp. 749--766, 2022.

\bibitem{Feng2020Privacy}
J.~Feng, L.~T. Yang, R.~Zhang, W.~Qiang, and J.~Chen, ``Privacy preserving
  high-order bi-lanczos in cloud-fog computing for industrial applications,''
  \emph{IEEE Trans. Ind. Inform.}, pp. 1--1, 2020.

\bibitem{Zhang2021Data}
H.~Zhang, C.~Tian, Y.~Li, L.~Su, N.~Yang, W.~X. Zhao, and J.~Gao, ``Data
  poisoning attack against recommender system using incomplete and perturbed
  data,'' in \emph{Proc. ACM SIGKDD Int. Conf. Knowl. Discov. Data Min.}, 2021,
  pp. 2154--2164.

\bibitem{Wang2021From}
Y.~Wang, Z.~Liu, L.~Y. Zhang, F.~Pareschi, G.~Setti, and G.~Chen, ``From chaos
  to pseudorandomness: A case study on the {2-D} coupled map lattice,''
  \emph{IEEE T. Cybern.}, pp. 1--11, 2021.

\bibitem{Zhang2023Buffeting}
Z.~Zhang, H.~Zhu, P.~Ban, Y.~Wang, and L.~Y. Zhang, ``Buffeting chaotification
  model for enhancing chaos and its hardware implementation,'' \emph{IEEE
  Trans. Ind. Electron.}, vol.~70, no.~3, pp. 2916--2926, 2023.

\bibitem{Zhang2020Novel}
Z.~Zhang, Y.~Wang, L.~Y. Zhang, and H.~Zhu, ``A novel chaotic map constructed
  by geometric operations and its application,'' \emph{Nonlinear Dyn.}, vol.
  102, no.~4, pp. 2843--2858, 2020.

\bibitem{Basilakis2021Efficient}
J.~Basilakis and B.~Javadi, ``Efficient parallel binary operations on
  homomorphic encrypted real numbers,'' \emph{IEEE Trans. Emerg. Top. Comput.},
  vol.~9, no.~1, pp. 507--519, 2021.

\bibitem{Badawi2021Towards}
A.~Al~Badawi, C.~Jin, J.~Lin, C.~F. Mun, S.~J. Jie, B.~H.~M. Tan, X.~Nan,
  K.~M.~M. Aung, and V.~R. Chandrasekhar, ``Towards the alexnet moment for
  homomorphic encryption: {HCNN}, the first homomorphic {CNN} on encrypted data
  with {GPUs},'' \emph{IEEE Trans. Emerg. Top. Comput.}, vol.~9, no.~3, pp.
  1330--1343, 2021.

\bibitem{Chen2021Privacy}
J.~Chen, K.~Li, and P.~S. Yu, ``Privacy-preserving deep learning model for
  decentralized {VANETs} using fully homomorphic encryption and blockchain,''
  \emph{IEEE Trans. Intell. Transp. Syst.}, pp. 1--10, 2021.

\bibitem{Ghimire2022Recent}
B.~Ghimire and D.~B. Rawat, ``Recent advances on federated learning for
  cybersecurity and cybersecurity for federated learning for internet of
  things,'' \emph{IEEE Internet Things J.}, pp. 1--1, 2022.

\bibitem{Gu2021Privacy}
B.~Gu, A.~Xu, Z.~Huo, C.~Deng, and H.~Huang, ``Privacy-preserving asynchronous
  vertical federated learning algorithms for multiparty collaborative
  learning,'' \emph{IEEE Trans. Neural Netw. Learn. Syst.}, pp. 1--13, 2021.

\bibitem{Yang2019Federated}
Q.~Yang, Y.~Liu, Y.~Cheng, Y.~Kang, T.~Chen, and H.~Yu, ``Federated learning,''
  \emph{Synthesis Lectures on Artificial Intelligence and Machine Learning},
  vol.~13, no.~3, pp. 1--207, 2019.

\bibitem{Wang2021Privacy}
D.~Wang, J.~Ren, Z.~Wang, X.~Pang, Y.~Zhang, and X.~S. Shen,
  ``Privacy-preserving streaming truth discovery in crowdsourcing with
  differential privacy,'' \emph{IEEE. Trans. Mob. Comput.}, pp. 1--1, 2021.

\bibitem{Zheng2021Decentralized}
Z.~Zheng, T.~Wang, A.~K. Bashir, M.~Alazab, S.~Mumtaz, and X.~Wang, ``A
  decentralized mechanism based on differential privacy for privacy-preserving
  computation in smart grid,'' \emph{IEEE Trans. Comput.}, pp. 1--1, 2021.

\bibitem{Ma2021Data}
C.~Ma, L.~Yuan, L.~Han, M.~Ding, R.~Bhaskar, and J.~Li, ``Data level privacy
  preserving: A stochastic perturbation approach based on differential
  privacy,'' \emph{IEEE Trans. Knowl. Data Eng.}, pp. 1--1, 2021.

\bibitem{Abadi2016Deep}
M.~Abadi, A.~Chu, I.~Goodfellow, H.~B. McMahan, I.~Mironov, K.~Talwar, and
  L.~Zhang, ``Deep learning with differential privacy,'' in \emph{Proceedings
  of the 2016 ACM SIGSAC conference on computer and communications security},
  2016, pp. 308--318.

\bibitem{Phan2017Adaptive}
N.~Phan, X.~Wu, H.~Hu, and D.~Dou, ``Adaptive {Laplace} mechanism: Differential
  privacy preservation in deep learning,'' in \emph{2017 IEEE International
  Conference on Data Mining (ICDM)}, 2017, pp. 385--394.

\bibitem{Wu2019P3SGD}
B.~Wu, S.~Zhao, G.~Sun, X.~Zhang, Z.~Su, C.~Zeng, and Z.~Liu, ``{P3SGD}:
  Patient privacy preserving {SGD} for regularizing deep {CNNs} in pathological
  image classification,'' in \emph{2019 IEEE/CVF Conference on Computer Vision
  and Pattern Recognition (CVPR)}, 2019, pp. 2094--2103.

\bibitem{Wang2020Deep}
Q.~Wang, Z.~Li, Q.~Zou, L.~Zhao, and S.~Wang, ``Deep domain adaptation with
  differential privacy,'' \emph{IEEE Trans. Inf. Forensic Secur.}, vol.~15, pp.
  3093--3106, 2020.

\bibitem{CHEN2021An}
X.~Chen, T.~Zhang, S.~Shen, T.~Zhu, and P.~Xiong, ``An optimized differential
  privacy scheme with reinforcement learning in {VANET},'' \emph{Comput.
  Secur.}, vol. 110, p. 102446, 2021.

\bibitem{Yang2010Nature}
X.-S. Yang, \emph{Nature-inspired metaheuristic algorithms}.\hskip 1em plus
  0.5em minus 0.4em\relax Luniver press, 2010.

\bibitem{Chou2018Forward}
J.-S. Chou and T.-K. Nguyen, ``Forward forecast of stock price using
  sliding-window metaheuristic-optimized machine-learning regression,''
  \emph{IEEE Trans. Ind. Inform.}, vol.~14, no.~7, pp. 3132--3142, Jul. 2018.

\bibitem{Zhao2021Overview}
S.~, F.~Blaabjerg, and H.~Wang, ``An overview of artificial intelligence
  applications for power electronics,'' \emph{IEEE Trans. Power Electron.},
  vol.~36, no.~4, pp. 4633--4658, Apr. 2021.

\bibitem{Fong2018How}
S.~Fong, S.~Deb, and X.-s. Yang, ``How meta-heuristic algorithms contribute to
  deep learning in the hype of big data analytics,'' in \emph{Progress in
  intelligent computing techniques: theory, practice, and applications}.\hskip
  1em plus 0.5em minus 0.4em\relax Springer, 2018, pp. 3--25.

\bibitem{Chen2021Novel}
D.~Chen, X.~Li, and S.~Li, ``A novel convolutional neural network model based
  on beetle antennae search optimization algorithm for computerized tomography
  diagnosis,'' \emph{IEEE Trans. Neural Netw. Learn. Syst.}, pp. 1--12, 2021.

\bibitem{mohana2017heuristics}
S.~Mohana and S.~S.~A. Mary, ``Heuristics for privacy preserving data mining:
  An evaluation,'' in \emph{2017 International Conference on Algorithms,
  Methodology, Models and Applications in Emerging Technologies
  (ICAMMAET)}.\hskip 1em plus 0.5em minus 0.4em\relax IEEE, 2017, pp. 1--9.

\bibitem{xie2021differential}
Y.~Xie, P.~Li, J.~Zhang, and M.~R. Ogiela, ``Differential privacy distributed
  learning under chaotic quantum particle swarm optimization,''
  \emph{Computing}, vol. 103, pp. 449--472, 2021.

\bibitem{Wang2016Secure}
C.~Wang, K.~Ren, and J.~Wang, ``Secure optimization computation outsourcing in
  cloud computing: A case study of linear programming,'' \emph{IEEE Trans.
  Comput.}, vol.~65, no.~1, pp. 216--229, 2016.

\bibitem{Zhan2021Evolutionary}
Z.-H. Zhan, S.-H. Wu, and J.~Zhang, ``A new evolutionary computation framework
  for privacy-preserving optimization,'' in \emph{2021 13th International
  Conference on Advanced Computational Intelligence (ICACI)}, 2021, pp.
  220--226.

\bibitem{Zhang2013Privgene}
J.~Zhang, X.~Xiao, Y.~Yang, Z.~Zhang, and M.~Winslett, ``Privgene:
  differentially private model fitting using genetic algorithms,'' in
  \emph{Proceedings of the 2013 ACM SIGMOD International Conference on
  Management of Data}, 2013, pp. 665--676.

\bibitem{Dwork2008Differential}
C.~Dwork, ``Differential privacy: A survey of results,'' in \emph{International
  conference on theory and applications of models of computation}.\hskip 1em
  plus 0.5em minus 0.4em\relax Springer, 2008, pp. 1--19.

\bibitem{Sweeney2002K}
L.~Sweeney, ``k-anonymity: A model for protecting privacy,''
  \emph{International Journal of Uncertainty, Fuzziness and Knowledge-Based
  Systems}, vol.~10, no.~05, pp. 557--570, 2002.

\bibitem{Dwork2014Algorithmic}
C.~Dwork, A.~Roth \emph{et~al.}, ``The algorithmic foundations of differential
  privacy.'' \emph{Found. Trends Theor. Comput. Sci.}, vol.~9, no. 3-4, pp.
  211--407, 2014.

\bibitem{Tang2021Review}
J.~Tang, G.~Liu, and Q.~Pan, ``A review on representative swarm intelligence
  algorithms for solving optimization problems: Applications and trends,''
  \emph{IEEE-CAA J. Automatica Sin.}, vol.~8, no.~10, pp. 1627--1643, 2021.

\bibitem{Kennedy1995Particle}
J.~Kennedy and R.~Eberhart, ``Particle swarm optimization,'' in
  \emph{Proceedings of ICNN'95-international conference on neural networks},
  vol.~4.\hskip 1em plus 0.5em minus 0.4em\relax IEEE, 1995, pp. 1942--1948.

\bibitem{Mmirjalili2014Grey}
S.~Mirjalili, S.~M. Mirjalili, and A.~Lewis, ``Grey wolf optimizer,''
  \emph{Adv. Eng. Softw.}, vol.~69, pp. 46--61, 2014.

\bibitem{Mirjalili2016Whale}
S.~Mirjalili and A.~Lewis, ``The whale optimization algorithm,'' \emph{Adv.
  Eng. Softw.}, vol.~95, pp. 51--67, 2016.

\bibitem{Dhiman2019Seagull}
G.~Dhiman and V.~Kumar, ``Seagull optimization algorithm: Theory and its
  applications for large-scale industrial engineering problems,''
  \emph{Knowledge-Based Syst.}, vol. 165, pp. 169--196, 2019.

\bibitem{Li2021When}
X.~Li, Y.~Chen, C.~Wang, and C.~Shen, ``When deep learning meets differential
  privacy: Privacy,security, and more,'' \emph{IEEE Netw.}, vol.~35, no.~6, pp.
  148--155, Nov. 2021.

\bibitem{zhao2022survey}
Y.~Zhao and J.~Chen, ``A survey on differential privacy for unstructured data
  content,'' \emph{ACM Computing Surveys (CSUR)}, vol.~54, no. 10s, pp. 1--28,
  2022.

\bibitem{Kennedy1997Particle}
J.~Kennedy, ``The particle swarm: social adaptation of knowledge,'' in
  \emph{Proceedings of 1997 IEEE International Conference on Evolutionary
  Computation (ICEC'97)}.\hskip 1em plus 0.5em minus 0.4em\relax IEEE, 1997,
  pp. 303--308.

\bibitem{Wang2013Diversity}
H.~Wang, H.~Sun, C.~Li, S.~Rahnamayan, and J.-s. Pan, ``Diversity enhanced
  particle swarm optimization with neighborhood search,'' \emph{Inf. Sci.},
  vol. 223, pp. 119--135, 2013.

\bibitem{Cavagna2022Marginal}
A.~Cavagna, A.~Culla, X.~Feng, I.~Giardina, T.~S. Grigera, W.~Kion-Crosby,
  S.~Melillo, G.~Pisegna, L.~Postiglione, and P.~Villegas, ``Marginal speed
  confinement resolves the conflict between correlation and control in
  collective behaviour,'' \emph{Nat. Commun.}, vol.~13, no.~1, pp. 1--11, 2022.

\bibitem{Little2000Millennial}
T.~D. Little, ``The millennial challenge: Modelling the agentic self in
  context,'' \emph{International Journal of Behavioral Development}, vol.~24,
  no.~2, pp. 149--152, 2000.

\bibitem{Gao2021Chaotic}
S.~Gao, Y.~Yu, Y.~Wang, J.~Wang, J.~Cheng, and M.~Zhou, ``Chaotic local
  search-based differential evolution algorithms for optimization,'' \emph{IEEE
  Trans. Syst. Man Cybern. -Syst.}, vol.~51, no.~6, pp. 3954--3967, Jun. 2021.

\bibitem{Xu2019Enhanced}
Y.~Xu, H.~Chen, J.~Luo, Q.~Zhang, S.~Jiao, and X.~Zhang, ``Enhanced moth-flame
  optimizer with mutation strategy for global optimization,'' \emph{Inf. Sci.},
  vol. 492, pp. 181--203, Aug. 2019.

\bibitem{Dwork2015Reusable}
C.~Dwork, V.~Feldman, M.~Hardt, T.~Pitassi, O.~Reingold, and A.~Roth, ``The
  reusable holdout: Preserving validity in adaptive data analysis,''
  \emph{Science}, vol. 349, no. 6248, pp. 636--638, Aug. 2015.

\bibitem{Zhao2020Privacy}
L.~Zhao, Q.~Wang, Q.~Zou, Y.~Zhang, and Y.~Chen, ``Privacy-preserving
  collaborative deep learning with unreliable participants,'' \emph{IEEE Trans.
  Inf. Forensic Secur.}, vol.~15, pp. 1486--1500, 2020.

\bibitem{Lykouris2016Learning}
T.~Lykouris, V.~Syrgkanis, and {\'E}.~Tardos, ``Learning and efficiency in
  games with dynamic population,'' in \emph{Proceedings of the twenty-seventh
  annual ACM-SIAM symposium on Discrete algorithms}.\hskip 1em plus 0.5em minus
  0.4em\relax SIAM, 2016, pp. 120--129.

\end{thebibliography}
\end{document}